%% file: main.tex
\documentclass[11pt]{article}

\usepackage[preprint]{acl}
\IfFileExists{latexml.sty}{\usepackage{latexml}}{\newif\iflatexml}
\hypersetup{
  hidelinks,
  pdftitle={NumTrace: Auditing the Number Provenance of Financial Analyst Agents},
  pdfauthor={Tianlei Zhu, Matt White, Xue Liu, Xiao-Yang Liu}
}

\usepackage{times}
\usepackage{latexsym}
\usepackage[T1]{fontenc}
\usepackage[utf8]{inputenc}
\usepackage{microtype}
\usepackage{inconsolata}
\usepackage{graphicx}
\usepackage{booktabs}
\usepackage{array}
\usepackage{colortbl}
\usepackage{enumitem}
\usepackage{multirow}
\usepackage{amsmath}
\usepackage{amssymb}
\usepackage{xcolor}
\usepackage[most]{tcolorbox}
\usepackage{pifont}
\colorlet{rnwarevision}{black}
\colorlet{v10rev}{black} 
\newcommand{\newrev}[1]{{\color{v10rev}\hypersetup{citecolor=v10rev,urlcolor=v10rev,linkcolor=v10rev}#1}}
\usepackage{tikz}

\makeatletter
\g@addto@macro{\UrlBreaks}{\do\-}
\makeatother

\title{NumTrace: Auditing the Number Provenance of Financial Analyst Agents}

\iflatexml
\author{
  Tianlei Zhu\\
  Columbia University\\
  \texttt{tz2617@columbia.edu}
  \And
  Matt White\\
  Columbia University\\
  University of California, Berkeley\\
  Linux Foundation\\
  \texttt{matt.white@berkeley.edu}
  \And
  Xue Liu\\
  MBZUAI\\
  McGill University\\
  Mila -- Quebec AI Institute\\
  \texttt{xue.liu@mcgill.ca}
  \And
  Xiao-Yang Liu\textsuperscript{*}\\
  Columbia University\\
  \texttt{xl2427@columbia.edu}
}
\else
\author{
  Tianlei Zhu\textsuperscript{1},\quad
  Matt White\textsuperscript{1,2,3},\quad
  Xue Liu\textsuperscript{4,5,6},\quad
  and Xiao-Yang Liu\textsuperscript{1}\thanks{Corresponding author: \texttt{xl2427@columbia.edu}.}\\
  \textsuperscript{1}Columbia University\quad
  \textsuperscript{2}University of California, Berkeley\quad
  \textsuperscript{3}Linux Foundation\\
  \textsuperscript{4}MBZUAI\quad
  \textsuperscript{5}McGill University\quad
  \textsuperscript{6}Mila -- Quebec AI Institute\\
  Emails: \texttt{\{tz2617, xl2427\}@columbia.edu},\\
  \texttt{matt.white@berkeley.edu},\quad
  \texttt{xue.liu@mcgill.ca}
}
\fi

\begin{document}
\maketitle

\input{sections/abstract}

\input{sections/01_intro}

\input{sections/02_benchmark}
\input{sections/03_method}
\input{sections/04_results}  

\input{sections/limitations}
\input{sections/ethics}

\makeatletter
\ifacl@finalcopy
\section*{Acknowledgments}

Tianlei Zhu and Xiao-Yang Liu acknowledge support from Columbia University's SEAS Interdisciplinary Research (SIRS) and SEAS Translational Acceleration Research (STAR) Seed Funding Program and the Tang Family Fund for Research Innovations in FinTech, Engineering, and Business Operations, and the Center for Digital Finance and Technologies (CDFT). Xiao-Yang Liu acknowledges support from the NSF IUCRC CRAFT Center (CRAFT Grant 22017) and the JPMorganChase Faculty Research Award.

The opinions expressed in this paper are those of the authors and do not necessarily reflect those of NSF IUCRC CRAFT, JPMorganChase, or its affiliates. This material is not a product of the Research Department of J.P. Morgan Securities LLC and does not constitute investment advice, a recommendation, or a solicitation or offer in any jurisdiction.
\fi
\makeatother

\bibliography{references}

\appendix
\input{sections/appendix}

\end{document}

%% file: sections/abstract.tex
\begin{abstract}
\label{sec:abstract}
A numerical claim generated by a financial analyst agent can match a number in a company filing but refer to the wrong entity, concept, period, unit or scope. A check based on reference answers can mark such an answer as wrong but does not explain why it fails. In this paper, we present NumTrace, a new benchmark that audits the provenance of each numerical claim made by financial analyst agents: its source and financial semantics. NumTrace currently has 540 questions over 2,163 SEC EDGAR documents. To the best of our knowledge, it is the first numerical benchmark about who holds a company's shares (Form~13F) and what its executives are owed (Item~402(i)). On a frozen snapshot of these documents, we build the reference answers according to accounting rules. NumTrace then gives each claim three separate verdicts: value, binding and support. {\color{rnwarevision}\hypersetup{citecolor=rnwarevision,urlcolor=rnwarevision,linkcolor=rnwarevision}By retaining separate value and binding verdicts, NumTrace detects \emph{right number, wrong account} (RNWA) errors: a claim has an RNWA error when it passes the value check but fails the binding check. The RNWA rate quantifies the gap between value-only and joint value-and-binding pass rates.} Finally, we audit three agents, Claude Opus~5.5, GPT-6 Astra and DeepSeek-V4.1-Flash. RNWA errors make up 67.5--81.8\% of each agent's failing claims.
In particular, on 13F co-filing group questions, about 91\% of claims pass the value check, but fewer than half pass the binding check.
\end{abstract}

%% file: sections/01_intro.tex
\section{Introduction}
\label{sec:intro}

\begin{figure}[t]
\centering
\includegraphics[width=\columnwidth]{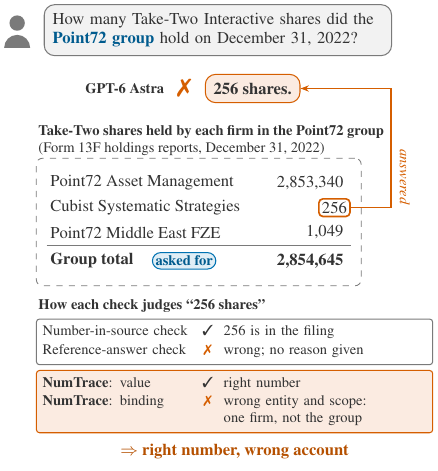}
{\captionsetup{labelfont={color=rnwarevision},textfont={color=rnwarevision}}
\hypersetup{citecolor=rnwarevision,urlcolor=rnwarevision,linkcolor=rnwarevision}
\caption{An RNWA error: GPT-6 Astra reports one firm's 256 shares as the Point72 group's holding (reference value: 2,854,645). Figure~\ref{fig:provenance-example} gives the full record.}\label{fig:teaser}}
\end{figure}

\begin{figure*}[t]
\centering
\includegraphics[width=\textwidth]{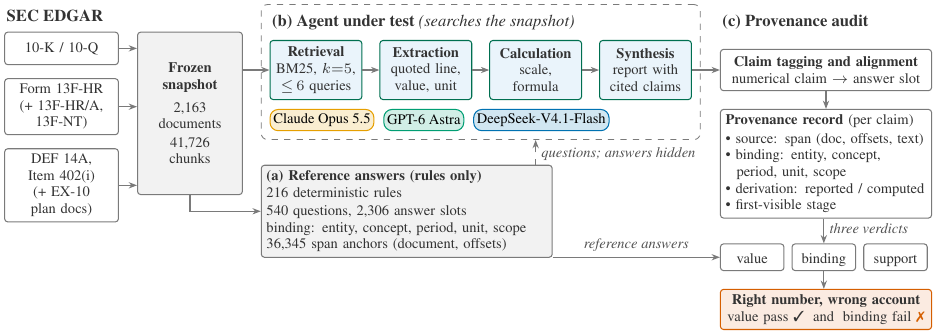}
\caption{Overview of the NumTrace benchmark. (a)~On a frozen snapshot of SEC filings, we build the questions and reference answers according to accounting rules. (b)~The agent under test gets the question and searches the snapshot; it never sees the reference answers. (c)~Each reported number gets a provenance record and three checks; a number that passes the value check but fails the binding check is a right number, wrong account (RNWA) error.}
\label{fig:overview}
\end{figure*}

FinGPT collects and curates financial data and uses parameter-efficient fine-tuning to adapt general-purpose large language models (LLMs) to financial tasks \citep{liu2023fingpt}.
Financial analyst agents powered by LLMs now read company filings in the database of the U.S. Securities and Exchange Commission (SEC) and report numbers for investment and compliance work \citep{tian2024fingptsearch,zhu2025findeepresearch,hu2025finsearchcomp}.
The survey of multimodal financial foundation models (MFFMs) identifies financial-data interpretation and output reliability as key challenges \citep{liu2025mffms}.
Consider this question: \textit{how many Take-Two Interactive shares did the Point72 group hold at the end of 2022?}
GPT-6 Astra answered ``256 shares,'' a number retrieved from a Form~13F holdings report, as shown in Figure~\ref{fig:teaser}.
The number is right, yet the answer is wrong: 256 is one firm's holding, while the total holding is 2,854,645 shares.
We call this type of error \emph{right number, wrong account} (RNWA).
Two common checks cannot diagnose this error.
A number-in-source check accepts it because 256 appears in the filing.
A reference-answer check compares the answer with the reference answer and rejects it, but it does not explain why it fails.

{\color{rnwarevision}\hypersetup{citecolor=rnwarevision,urlcolor=rnwarevision,linkcolor=rnwarevision}Such errors can arise because accounting rules determine what a reported number means, even when those rules are not printed next to it.}
{\color{rnwarevision}\hypersetup{citecolor=rnwarevision,urlcolor=rnwarevision,linkcolor=rnwarevision}
The Form~13F instructions set the units by filing date and define which affiliated managers count toward a group total \citep{sec2023form13f}; Item~402(i) separates deferred compensation by plan \citep{cfr229402}.
Evaluating provenance therefore requires checking both source and meaning \citep{hall2026verifin} (\S\ref{sec:benchmark-filings}).
\citet{liu2026certified} argue that in policy-governed domains an action can be computable yet impermissible, so execution should wait for a certified trace.
NumTrace gives this idea a concrete, measurable financial setting: an RNWA claim is computable from the filing yet violates the rules that fix the number's meaning.%
}

{
XBRL associates reported numbers with accounting concepts, entities, periods and units \citep{han2024xbrlagent,wang2025fintagging}.
NumTrace uses these facts and their inline XBRL spans to establish reference bindings for statement claims.
The audit checks whether an agent's answer preserves both the value and its meaning.
We apply the same checks to Form~13F and Item~402(i), with filing-specific rules supplying the semantics that the available tags do not capture.%
}

In this paper, we present NumTrace, a new benchmark that audits the provenance of each numerical claim, as shown in Figure~\ref{fig:overview}. Compared with existing evaluations, it adds three things:
\begin{itemize}[leftmargin=*,itemsep=1pt,topsep=2pt]
\item \textbf{New questions.} Of the 540 questions, 420 ask who holds a company's shares (Form~13F) or what its executives are owed (Item~402(i)). The closest benchmarks (Table~\ref{tab:positioning}) do not ask such questions (\S\ref{sec:benchmark}; Figure~\ref{fig:composition}).
\item \textbf{Checkable reference answers.} Written rules derive the reference answers and their source spans from a frozen snapshot of 2,163 SEC EDGAR documents; a programmer who re-implemented the rules independently obtained the same values for 443 of 444 sampled answer slots (\S\ref{sec:validation}).
\item {\color{rnwarevision}\hypersetup{citecolor=rnwarevision,urlcolor=rnwarevision,linkcolor=rnwarevision}\textbf{A diagnostic error measure.} NumTrace evaluates value, binding and support separately. Its RNWA rate, the gap between value-only and joint value-and-binding pass rates, isolates semantic errors that pass the value check (\S\ref{sec:method-measures}).}
\end{itemize}

{\color{rnwarevision}\hypersetup{citecolor=rnwarevision,urlcolor=rnwarevision,linkcolor=rnwarevision}Related evaluations assess answer correctness, evidence support or the consistency of structured claims.}
Financial question answering (QA) benchmarks score final answers over 10-K and 10-Q reports \citep{chen2021finqa,zhu2021tatqa,reddy2024docfinqa,islam2023financebench}, where the meaning of a number is mostly printed next to the number.
Methods for claim verification, attribution and evidence ranking \citep{zhao2024findver,gao2023alce,mansouri2026finrank} judge whole answers or give one label per claim.
The closest work does check what a number refers to.
FinGround matches the value, unit, period and entity of atomic claims \citep{guo2026finground}, and VeriFin grounds answer operands in XBRL facts \citep{hall2026verifin}.
FinFIRST checks entities, periods, units and definitions with rubrics \citep{wang2026finfirst}, and BigFinanceBench checks entities, periods and concepts \citep{wang2026bigfinancebench}.
These checks merge a number's value and its meaning into one label or rubric item.
{\color{rnwarevision}\hypersetup{citecolor=rnwarevision,urlcolor=rnwarevision,linkcolor=rnwarevision}NumTrace keeps the joint value--binding outcome of each claim, so value-pass, binding-fail errors can be measured on their own (Table~\ref{tab:compare}; Appendix~\ref{sec:related}).}

\input{tables/compare}

{\color{rnwarevision}\hypersetup{citecolor=rnwarevision,urlcolor=rnwarevision,linkcolor=rnwarevision}
Each numerical claim receives a \emph{provenance record} (\S\ref{sec:method-record}) and three deterministic checks: \emph{Value} (digits and arithmetic match the source), \emph{Binding} (the number refers to what the claim states) and \emph{Support} (an authoritative source backs both; \S\ref{sec:method-verdicts}).%
}

{\color{rnwarevision}\hypersetup{citecolor=rnwarevision,urlcolor=rnwarevision,linkcolor=rnwarevision}
We audit Claude Opus~5.5, GPT-6 Astra and DeepSeek-V4.1-Flash in a shared four-stage pipeline (\S\ref{sec:setup}).
RNWA errors account for 67.5--81.8\% of failing claims, with co-filing groups the hardest family (Table~\ref{tab:rnwa-share}; \S\ref{sec:analysis}).
An expert review of 360 sampled claims supports this pattern (\S\ref{sec:validation}).%
}

%% file: tables/compare.tex
\begin{table}[t]
\centering\footnotesize
\newsavebox{\ntcy}\sbox{\ntcy}{\tikz[baseline=-0.6ex]\fill (0,0) circle (2.4pt);}%
\newcommand{\cy}{\usebox{\ntcy}}%
\newsavebox{\ntcp}\sbox{\ntcp}{\tikz[baseline=-0.6ex]{\draw[line width=0.5pt] (0,0) circle (2.15pt); \fill (0,2.4pt) arc (90:270:2.4pt) -- cycle;}}%
\newcommand{\cp}{\usebox{\ntcp}}%
\newsavebox{\ntcn}\sbox{\ntcn}{\tikz[baseline=-0.6ex]\draw[line width=0.5pt] (0,0) circle (2.15pt);}%
\newcommand{\cn}{\usebox{\ntcn}}%
\newcommand{\cnr}{\textcolor{black!55}{\textbf{?}}}%
\definecolor{ntRow}{HTML}{FBEBE1}
\setlength{\tabcolsep}{2.6pt}
\begin{tabular}{@{}lccccc@{}}
\toprule
 & {\scriptsize Per} & {\scriptsize Checks} & {\scriptsize Value,} & {\scriptsize Error} & {\scriptsize 13F/} \\
Work & {\scriptsize claim} & {\scriptsize meaning} & {\scriptsize binding apart} & {\scriptsize stage} & {\scriptsize 402(i)} \\
\midrule
FinRank & \cn & \cn & \cn & \cn & \cn \\
Fin-RATE & \cn & \cy & \cn & \cy & \cn \\
VeriFin & \cy & \cy & \cn & \cnr & \cn \\
FinGround & \cy & \cy & \cn & \cp & \cnr \\
FinFIRST & \cn & \cy & \cn & \cp & \cnr \\
BigFinanceBench & \cn & \cy & \cn & \cy & \textcolor{v10rev}{\cn} \\
FinTrace & \cn & \cn & \cn & \cp & \cn \\
\midrule
\rowcolor{ntRow}\textbf{NumTrace} & \cy & \cy & \cy & \cy & \cy \\
\bottomrule
\end{tabular}
\caption{NumTrace and the closest work, checked against each paper. \cy\ yes, \cp\ partly, \cn\ no, \cnr\ not stated in the paper. The other works that check a number's meaning merge it with the value into one label or rubric item. Details: Appendix~\ref{sec:related}.}
\label{tab:compare}
\end{table}

%% file: sections/02_benchmark.tex
\section{The NumTrace Benchmark}
\label{sec:benchmark}

NumTrace has 540 questions in 12 families over three filing types (Figure~\ref{fig:composition}; datasheet in Appendix~\ref{app:datasheet}).
The questions have 2,306 answer slots in total; a slot is one field of a reference answer.
The questions ask for 1,579 of these slots directly; 527 are inputs to computed slots, and 200 are kept only for documentation.

\subsection{A Frozen Snapshot}
\label{sec:benchmark-snapshot}

Filings change over time, so all evidence comes from one frozen snapshot of SEC EDGAR.
The snapshot holds 2,163 documents (2,132 filings and 31 plan documents), cut into 41,726 chunks.
Its as-of dates run from 2022Q3 to 2026Q2.
A SHA-256 hash of the ordered document list identifies the snapshot, and the loader checks it.
The renderer writes each table cell as \textit{row label $|$ column header: number} and keeps each table in one chunk; it writes 13F holdings one row per line, without a unit.

\subsection{Filing Types and Traps}
\label{sec:benchmark-filings}

\begin{figure}[t]
\centering
\includegraphics[width=\columnwidth]{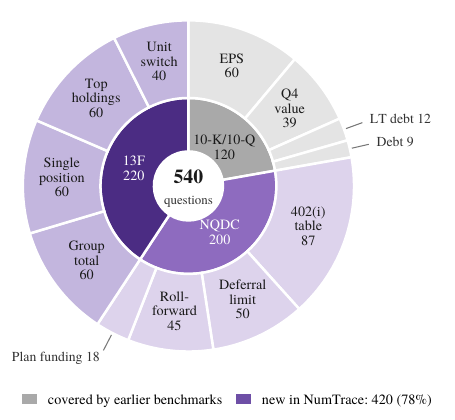}
\caption{What NumTrace asks about. Of the 540 questions, 420 (purple) ask about 13F holdings and Item~402(i) deferred-compensation tables, which the closest benchmarks (Table~\ref{tab:positioning}) do not ask about.}
\label{fig:composition}
\end{figure}

\paragraph{Financial statements (120 questions).}
{\color{rnwarevision}\hypersetup{citecolor=rnwarevision,urlcolor=rnwarevision,linkcolor=rnwarevision}
Statements provide the baseline: questions cover debt, diluted earnings per share (EPS) and fourth-quarter values in XBRL-tagged 10-K and 10-Q filings from 30 large U.S. companies.%
}

\paragraph{Item 402(i) deferred compensation (200 questions).}
Here an agent must tie each balance to the right plan, officer and year.
For the same 30 companies, the questions target the nonqualified deferred compensation (NQDC) table of the proxy statement, the annual filing that discloses executive pay \citep{cfr229402}.
{\color{rnwarevision}\hypersetup{citecolor=rnwarevision,urlcolor=rnwarevision,linkcolor=rnwarevision}
Questions also use the 10-K and the plan documents.
}
The table reports each plan separately \citep{seccdi12503}.
When a column has nothing to report, Item~402(a)(5) lets the company omit it; the reference answer is then \textit{not reported}, and a correct answer says so.
Plan documents exist in several versions, and the reference answers use the version in force at the end of the fiscal year.
One family asks whether a plan is funded, a basic NQDC distinction \citep{irs2024nqdcatg}.
Sixteen questions are negative controls: their proxies have no 402(i) table.

{
The NQDC-FUND family also asks for company-wide plan liabilities and the cash surrender value of company-owned life insurance, using \texttt{us-gaap} facts from the fiscal year's 10-K.
Of these 36 slots, 10 have reported values and 26 are \textit{not found by rule}.
These questions combine plan documents with tagged 10-K facts where available.%
}

\paragraph{Form 13F holdings (220 questions).}
Here the traps are unit, amendment and filer set.
A documented selection rule chose 30 institutional managers from 111 candidates; with their co-filing group members, the corpus covers 92 managers.
Questions ask for single positions, top holdings, unit changes and group totals (Figure~\ref{fig:teaser} shows a group question).
The form's instructions create three traps.
First, the unit depends on the filing date: values are in thousands of dollars before 2023-01-03 and in dollars afterward \citep[FAQ~62]{sec13ffaq}.
Second, amendments differ: a restatement replaces earlier rows, while a new-holdings amendment adds rows \citep{sec2023form13f}.
Third, a group total depends on membership. Column~7 of each row names the other managers that share the position, and a group total counts a member's rows only if these names resolve to the group filer or to a manager on the filer's list \citep{sec2023form13f}.
13F-GROUP draws on 9 usable multi-member groups; in 24 of its 60 questions, no other member counts in that quarter, so the reference answer equals the filer's own rows.
We use the 13F filings, not the SEC's flat data sets \citep{sec13fdatasets}, so every number has a source span.

\subsection{Slots, Bindings and Reference Answers}
\label{sec:benchmark-gold}

Every slot records what its number refers to.
Each slot has a binding tuple (entity, concept, period, unit, scope).
For example, one 13F slot binds Oaktree Capital Management, the value of its position in the security with CUSIP G89479102 (a security identifier), 2024Q4, USD and scope \textit{filer}.
Scope separates a filer from a co-filing group in 13F, and an executive, plan or company in NQDC.

{
The binding fields align with the XBRL fact model: entity, concept, period and unit identify the reported quantity.
NumTrace adds scope to distinguish aggregation and reporting boundaries (Table~\ref{tab:xbrl-binding}).
Scope can serve a role analogous to a dimensional qualifier but is not implemented as an XBRL dimension.
Statement facts use \texttt{us-gaap} concepts; \texttt{13f:} and \texttt{nqdc:} are project concept prefixes, not formal XBRL taxonomy extensions.
Form~13F information tables use XML, while the SEC's ECD tagging requirements cover specified compensation and trading disclosures rather than Item~402(i) \citep{sec2026ecd}.%
}

Each slot also has a status: 1,391 are \textit{reported} and 661 are \textit{derived} by a stated formula; the rest are \textit{text} (138), \textit{not reported} (84), \textit{not found by rule} (26, no reference value) or \textit{not applicable} (6).

Deterministic Python rules extract the reference answers from the filings.
We listed 216 rules that affect the reference answers: 22 follow an external authority such as the Form~13F instructions or Item~402 (class A), and 194 are our own design choices (class B), where another reasonable choice would give different values for the slots it affects (Appendix~\ref{app:rules}).
An author confirmed every rule choice.
For printed values, 36,345 anchors from 21,962 facts give the document and character offsets. Of the 1,331 reported slots that are numeric, 11 have no anchor (Appendix~\ref{app:release}).
A clean rebuild reproduces every data file with identical MD5 hashes.
The question text contains no reference value and no accession number (the EDGAR filing identifier).
Beyond the question text, the prompt gives no concept, period, unit or scope, as these form the binding we evaluate (Appendix~\ref{app:prompts}).

{
\paragraph{Statement reference answers from XBRL.}
We build reference answers for reported statement items from the SEC Company Facts API and retain only facts filed on or before the snapshot freeze date, 2026-10-02.
The API aggregates standard-taxonomy facts that apply to the entire filing entity \citep{sec2025edgarapis}.
Of the 420 statement slots, 314 are \textit{reported}, 104 are \textit{derived}, and 2 are \textit{not reported}.
Derived values need not appear as individual tagged facts.
To anchor each fact in the filing text, we match its concept, reporting period and value to an inline XBRL span whose context has no explicit or typed dimensions.
Each anchor retains the printed number and scale.
The corpus contains 810,911 extracted tagged spans and \newrev{23,063} anchors linking statement facts to source spans.

When comparing values across filings, we use inline XBRL precision where available and otherwise fall back to the printed scale.
Some debt reference rows also retain related tagged facts as metadata.
The XOM example in Table~\ref{tab:binddims} shows why \texttt{DebtCurrent} and \texttt{LongTermDebtCurrent} must remain distinct.
Derived answers retain their input facts and formulas.
Metric definitions, derivations and source versions follow explicit rules.%
}

\subsection{Agents under Test}
\label{sec:setup}

\paragraph{Pipeline.}
Each agent pairs one model with the same fixed four-stage pipeline: retrieve, extract, calculate and synthesize.
Each stage logs structured output.
The model writes up to six queries, and each query returns its top $k{=}5$ chunks under Okapi BM25 ranking \citep{robertson2009bm25}.
The model then quotes each line it uses, picks a scale for each fact and a formula for each slot, and writes a short report with citations.
The harness runs retrieval and evaluates the formulas; the models have no other tools and no web access.

\paragraph{Agents.}
We test three agents on all 540 questions: Claude Opus~5.5 through the Claude Code command-line interface (CLI), GPT-6 Astra through the Codex CLI and DeepSeek-V4.1-Flash \citep{deepseek2026v41flash} through its application programming interface (API).
None used tools; versions and settings are in Appendix~\ref{app:prompts}, and tables use the short names Claude, GPT and DeepSeek.
Two agents call their models through coding-agent CLIs with all tools disabled, and one through an API, so the comparison is not like-for-like across base models.
Each agent has one scored run per question and returns a report for every question.

\paragraph{Retrieval coverage.}
Results depend on how much the agents can retrieve.
In an offline study with templated queries, six queries of five chunks each retrieved all needed chunks for 57\% of the 512 questions with scorable evidence, but for only 10\% of 13F-TOP questions, which ask for a manager's three largest positions.
Missing evidence therefore accounts for part of the 13F errors.

%% file: sections/03_method.tex
\section{Evaluation Method}
\label{sec:framework}

We score every numerical claim in a report, not only the final answer.
A tagger extracts claims and aligns them to slots, and each claim gets a provenance record. The deterministic scorer then assigns three verdicts against the reference answers, which we summarize per claim, slot and report.
The agent never sees reference values; only the scorer uses them (Appendix~\ref{app:scorer}).

\subsection{Claims and Provenance Records}
\label{sec:method-record}

A deterministic regular-expression tagger extracts the numerical claims of a report.
It aligns each claim to a slot by its citation, then by stated concept and period.
For every claim we record five fields:
(i) the \emph{source}, an exact span (document and offsets) in the snapshot;
(ii) the \emph{stated binding} (entity, concept, period, unit, scope), including the scale the agent applied to the printed number;
(iii) the \emph{reference binding} of the number printed at that source;
(iv) the \emph{derivation}, either reported (one copied input) or computed (a formula over logged inputs, each with its own span);
and (v) the \emph{first-visible stage}, the first logged stage whose output disagrees with the reference answer.
The record is checkable certification material \citep{liu2026certified}: it holds the evidence span, the stated binding and a replayable derivation.

\subsection{Three Verdicts and RNWA Errors}
\label{sec:method-verdicts}

Each claim gets three verdicts, each of which is pass, fail or not applicable.
\emph{Value} asks whether the digits and arithmetic match the source.
\emph{Binding} asks whether the number refers to what the claim says.
\emph{Support} asks whether the cited source is authoritative (a primary filing, not superseded) and backs both the number and its binding.
In Figure~\ref{fig:teaser}, the 256 shares are copied exactly, so value passes; they belong to one member filer, not to the Point72 group, so binding fails on entity and scope; and the cited row does not back a group holding, so support fails.

\paragraph{Value.}
Value is a defined check, not full numeric correctness.
For a reported claim, it asks whether the claim's digits equal the cited number's printed digits at the rounding the claim states.
Scale is left out: a claim that equals the printed number times a power of 1,000 passes, so ``\$0.54 million'' for \$542 printed in dollars passes the value check and fails the binding check on unit.
For a computed claim, each input's digits must match its source, and the formula over the agent's own inputs must give the claim at its rounding.
Signs are not compared with the reference value: filings print outflows as negative numbers, and a direction in words (``fell by'') only has to agree with the agent's own inputs.

\begin{table}[t]
\begin{minipage}[t]{\columnwidth}
\centering\footnotesize
\setlength{\tabcolsep}{3pt}
\begin{tabular}{@{}lp{0.76\linewidth}@{}}
\toprule
Dim. & Example of a binding failure (value pass) \\
\midrule
Entity, scope & 256 shares from one member filer's row reported as the Point72 group's holding (reference value 2,854,645 over three filers; GPT). \\
Concept & ``Long-term debt due within one year'' 4,892 reported for current debt 4,955 (XOM; Claude). \\
Period & Current portion of long-term debt 7,832 from the first-quarter 10-Q reported for the end of fiscal 2026 (reference value 9,227; MSFT; DeepSeek). \\
Unit & A 402(i) column printed in dollars read as thousands: ``\$0.54 million'' for \$542 (JPM; Claude). \\
\bottomrule
\end{tabular}
\caption{Binding dimensions (entity and scope share one example), with one real claim each from our runs. All four were confirmed by the expert review.}
\label{tab:binddims}
\end{minipage}
\end{table}

\paragraph{Binding.}
For a reported claim, we compare the stated binding with the reference binding of the cited number on all five dimensions. Table~\ref{tab:binddims} gives one real failure for each.
The stated binding takes entity, period and scale from the claim and the concept and scope from the aligned slot or from the quantity the report names.
When the source is anchored, the slot's reference value plays no role.
For a computed claim, we compare each input with the matching input of the slot (concept, period, scope); an input with the wrong scale fails on unit, and the result is compared with the slot's reference magnitude.
If the inputs are copied and the arithmetic is correct but the result differs from the reference value, the binding check fails on concept because the inputs or the operation do not match the slot.

\paragraph{Support.}
As in entailment-based attribution \citep{gao2023alce,yue2023attrscore}, the passage must support the claim; for numbers we make it explicit: support fails when the passage holds the number under another entity, period, unit or scope.

\begin{figure}[t]
\centering
\includegraphics[width=\columnwidth]{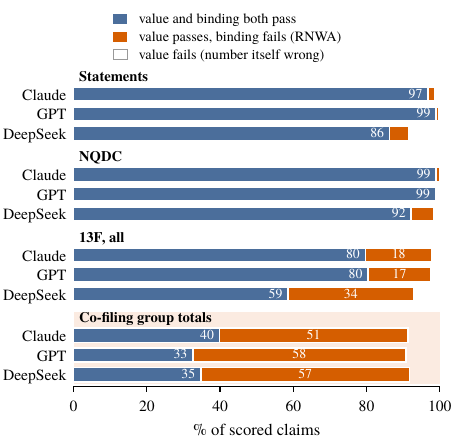}
\caption{Agents rarely miscopy numbers, but when asked for a co-filing group total, more than half of their scored claims pass the value check but fail the binding check. Each bar splits one agent's scored claims; co-filing group totals are a subset of 13F.}
\label{fig:headline}
\end{figure}

\paragraph{RNWA errors.}
A claim has an RNWA error when it passes the value check and fails the binding check.
For a reported claim, this means a correctly copied number whose entity, concept, period, unit or scope differs from the one the claim states.
For a computed claim, it also covers correctly executed arithmetic on the wrong inputs or operation, recorded as a concept failure; such a claim is a wrong answer, not always a mislabeled number.
A claim is \emph{scored} when it has a value verdict; it stays unscored when its cited source holds no anchored fact and it aligns to no slot with a reference value.
In Figure~\ref{fig:headline} and Table~\ref{tab:results}, every rate, the RNWA rate included, is a share of the scored claims. A \emph{failing} claim fails any check (examples in Appendix~\ref{app:traces}).

\subsection{Measures}
\label{sec:method-measures}

{\color{rnwarevision}\hypersetup{citecolor=rnwarevision,urlcolor=rnwarevision,linkcolor=rnwarevision}
\paragraph{The RNWA rate as a joint error rate.}
Let $\mathcal{S}$ be the set of scored claims, that is, the numbers in an agent's report that can be checked against a filing, and let $N=|\mathcal{S}|$.
For claim $i$, $V_i=1$ if the stated number matches the number printed in the cited filing (digits and arithmetic), and $B_i=1$ if the claim assigns the number to the right filer or group, concept, period, unit and scope; otherwise each is 0.
Both verdicts are binary for every scored claim in our evaluation; not-applicable verdicts are not treated as failures.
The empirical RNWA rate is
\begin{equation}
\begin{aligned}
\widehat r_{\mathrm{RNWA}}
&=\frac{\bigl|\{i\in\mathcal{S}: V_i=1,\ B_i=0\}\bigr|}{N}\\
&=\frac{1}{N}\sum_{i\in\mathcal{S}}V_i(1-B_i).
\end{aligned}
\label{eq:rnwa-rate}
\end{equation}
The term $V_i(1-B_i)$ equals 1 only when $V_i=1$ and $B_i=0$.
In practice, $\widehat r_{\mathrm{RNWA}}$ is the share of reported numbers that a reviewer who ties each number to the filing would accept, although they describe the wrong account, such as one member filer's 256 shares reported as the Point72 group's holding.
The RNWA rate is thus the gap between value-only and joint value-and-binding pass rates (Appendix~\ref{app:rnwa-marginals}).
In the terms of \citet{liu2026certified}, it is the certified-but-impermissible risk of a certifier that checks values only.
This concerns the two defined checks, not full financial correctness; support is assessed separately.
Tables~\ref{tab:results} and~\ref{tab:main} report $\widehat r_{\mathrm{RNWA}}$ over all scored claims; claim and slot coverage are reported separately.
Appendix~\ref{app:rnwa-marginals} explains why marginal pass rates do not determine the RNWA rate and gives its conditional-risk interpretation.%
}

\paragraph{Slots.}
Each asked slot is \emph{answered}, \emph{missing} (no number and no refusal) or \emph{refused} (an explicit non-answer).
A refusal is correct when the reference answer is \textit{not reported}.

\paragraph{First-visible stage.}
For a failing claim we report the first stage whose logged output disagrees with the reference answer.
We check extraction on the bound concept, period and unit text, calculation on the applied scale and slot, and synthesis on what the report states.
This heuristic label shows where an error first appears, not which stage causes it.

\paragraph{Reference mode and statistics.}
The default \emph{lenient} reference answer also accepts a slot's listed alternatives and a computed sum of parts for a printed total; the \emph{strict} one accepts only the primary value.
We report Wilson 95\% intervals over claims.
They ignore clustering by question, so for the key contrasts we add bootstrap intervals over questions or groups.
The frozen scorer reproduces all 36 saved scoring outputs byte for byte.

\subsection{Quality Control and Human Validation}
\label{sec:validation}

Both the reference answers and the scorer are error-prone, so we checked each.

\paragraph{Reference answers: domain review.}
A domain reviewer, a financial analyst with eight years of equity research experience, checked a seeded random sample of 200 questions (853 slots).
The sample came from an early version, before the fixes the review triggered.
Slot accuracy was 96.1\% [94.6, 97.2] (820 of 853), and 99.2\% on value alone.
All findings were fixed and logged, so these results describe the reviewed version rather than the released data.
An automated pre-check of an earlier version of the reference answers was used only to locate issues; we do not count it as validation.

\paragraph{Reference answers: independent re-implementation.}
A programmer with no access to the NumTrace code or reference answers implemented the rules from the written specification.
On 110 sampled questions, 443 of 444 slots agree under both lenient and strict reference answers.
The one disagreement is the spelling of a member name in a text slot.
This checks that the code follows the written rules, not that the rules read the filings correctly; we have no inter-annotator agreement.

\paragraph{Scorer debugging.}
We first debugged the scorer on a sample of Claude's failing claims. This review found 18 artifact classes (e.g., free-text periods, refusals scored as missing), and we fixed each in general code with tests; with no change to model outputs, reference answers or corpus, Claude's binding pass rate rose from 70.8\% to 93.8\% (Appendix~\ref{app:audit}).

\paragraph{Expert review of agent claims.}
After freezing the scorer, we drew a stratified sample of 420 items (360 claims and 60 unanswered slots): for each agent, 60 claims that passed every verdict, 60 claims that failed at least one and 20 unanswered slots.
Two LLM annotators, one GPT-based and one Claude-based, labeled every item separately.
A domain expert, a financial analyst with seven years of equity research experience, then judged every item against the EDGAR filing; the expert saw both LLM labels, but every final label is the expert's.
The review finds scorer errors in both directions (Table~\ref{tab:audit-claims}). Weighted to all scored claims, the binding pass rate is 88.0\% under the expert's labels against 90.6\% under the scorer, and DeepSeek stays lowest (80.7\%; Table~\ref{tab:audit-claims}).

%% file: sections/04_results.tex
\section{Results}
\label{sec:results}

All numbers here come from the automatic scorer; \S\ref{sec:validation} compares them with expert review.

\input{tables/results_main.tex}

\subsection{Value Checks Pass, Accounts Often Do Not}
\label{sec:results-master}

Value checks almost always pass (Figure~\ref{fig:headline}; Table~\ref{tab:results}): every agent passes the value check for at least 91.6\% of scored claims in every filing type.
Binding pass rates are lower, above all on 13F (65.4--82.7\%), and support pass rates follow them.

\begin{table}[t]
\centering\small
\setlength{\tabcolsep}{3.5pt}
\newlength{\nwA}\settowidth{\nwA}{90/110}\newcommand{\nA}[1]{\makebox[\nwA][r]{#1}}%
\newlength{\nwB}\settowidth{\nwB}{71/92}\newcommand{\nB}[1]{\makebox[\nwB][r]{#1}}%
\newlength{\nwC}\settowidth{\nwC}{233/345}\newcommand{\nC}[1]{\makebox[\nwC][r]{#1}}%
\begin{tabular}{@{}lccc@{}}
\toprule
 & Claude & GPT & DeepSeek \\
\midrule
RNWA share of failing claims & \nA{82\%} & \nB{77\%} & \nC{68\%} \\
RNWA / failing & \nA{90/110} & \nB{71/92} & \nC{233/345} \\
Share excl. result mismatches & \nA{72\%} & \nB{62\%} & \nC{54\%} \\
Result mismatches removed & \nA{39} & \nB{37} & \nC{103} \\
\bottomrule
\end{tabular}
\caption{Most failing claims have RNWA errors. The ``excl.'' row excludes result mismatches: RNWA claims whose only binding failure is a concept failure assigned because a correctly computed result differs from the reference value. A failing claim fails any check.}
\label{tab:rnwa-share}
\end{table}

Most failing claims have RNWA errors (Table~\ref{tab:rnwa-share}).
Part of this comes from our design: scale errors count as binding errors, and correct arithmetic with a result that differs from the reference value counts as a concept error.
Without the latter, RNWA errors are still the largest share of the remaining failing claims. In the expert review (\S\ref{sec:validation}), 86\% of sampled failing claims (131 of 152) have RNWA errors.
Concept is the most common failing binding dimension for every agent.
On 13F binding, a paired bootstrap over questions cannot separate Claude from GPT and places DeepSeek 10--22 points lower.

\subsection{Every Agent Fails on Co-Filing Groups}
\label{sec:analysis}

13F-GROUP asks for the combined holding of a manager and its affiliated co-filers, and it is the hardest family for every agent (Figure~\ref{fig:family-binding}): binding passes for only 41.9--48.5\% of claims, against 87.8--100\% for 13F-POS, which asks about one filer's position on the same form.
With only 9 usable groups, these rates are uncertain (Appendix~\ref{app:results}). The expert review points the same way. Of 14 sampled group claims that the scorer passed on binding, the expert judged 11 wrong; of 59 that it failed, the expert judged only 1 correct. The scorer, if anything, overstates these rates.

A common error is to report one filer's holding as the group total.
The family has 11\% of the questions but 44\% of all RNWA claims (172 of 394), and half of these (86) equal one filer's own position.
For example, Claude reports Jennison Associates' 23,240,649 Broadcom shares as Prudential's group holding (reference value 28,526,203).

\subsection{Interpretation for Evaluation Design}
\label{sec:results-design}

\paragraph{Score the account, not only the number.}
{\color{rnwarevision}\hypersetup{citecolor=rnwarevision,urlcolor=rnwarevision,linkcolor=rnwarevision}
Among RNWA claims with a numeric reference value (90, 71 and 213), reference-value comparison rejects 85, 69 and 183 but cannot distinguish copying, scale or binding failures.
The remaining 5, 2 and 30 equal the reference value, mostly NQDC claims with period errors.
The RNWA rate (Equation~\ref{eq:rnwa-rate}) quantifies the gap between value-only and joint value-and-binding pass rates.
As value pass rates approach one, the RNWA rate approaches the binding failure rate; its diagnostic role is to isolate errors that survive value checking.%
}

{
\paragraph{Use structured financial semantics.}
XBRL provides concept, period and unit evidence for statement reference answers; 13F and NQDC also require filing-specific rules.
Binding pass rates are higher on statements (92.5--99.3\%) than on 13F (65.4--82.7\%).
Yet NQDC tasks, which rely primarily on untagged proxy tables and plan documents, pass at rates similar to those on statements (93.7--99.7\%).
These tasks differ in complexity and retrieval coverage, so the comparison does not isolate an effect of XBRL.
Agents in our pipeline read rendered text without XBRL tags.
We have not tested whether those tags would help.%
}

\paragraph{Report coverage next to accuracy.}
Pass rates depend on what an agent chooses to state and on which claims the scorer scores.
Claude and GPT refuse 17.4\% and 17.8\% of asked slots, and DeepSeek refuses 3.3\%.
The share of written claims left unscored also differs (36.8\%, 4.3\% and 20.8\%), mostly because the cited source holds no anchored fact.
We therefore report coverage next to accuracy and do not rank the agents.

\paragraph{Debug the scorer.}
{\color{rnwarevision}\hypersetup{citecolor=rnwarevision,urlcolor=rnwarevision,linkcolor=rnwarevision}
Debugging increased Claude's binding pass rate from 70.8\% to 93.8\% with model outputs fixed. Scorers therefore need human validation.
Strict reference answers leave value pass rates unchanged and moves binding pass rates by at most 0.5 points and the RNWA count by at most 3 claims (Table~\ref{tab:strict}).%
}

\section{Conclusion}
\label{sec:conclusion}

{\color{rnwarevision}\hypersetup{citecolor=rnwarevision,urlcolor=rnwarevision,linkcolor=rnwarevision}
NumTrace audits numerical provenance in statements, Form~13F and Item~402(i).
Across three agents and an expert-reviewed sample, RNWA errors make up most failing claims, with co-filing groups the hardest family.
Exact rates depend on the scorer; group results rest on 9 usable groups.%
}

{\color{rnwarevision}\hypersetup{citecolor=rnwarevision,urlcolor=rnwarevision,linkcolor=rnwarevision}
\newrev{The RNWA rate separates value fidelity from binding correctness:} it is the loss in pass rate when a binding check is added to the value check.
XBRL facts and filing-specific rules make those bindings auditable across the three filing types.%
}

\paragraph{Outlook.}
NumTrace identifies and measures failures that a numerical-claim certifier would need to prevent. The next step is what an agent must prove before a number is released.
Following \citet{liu2026certified}, a runtime certifier would release a number only if its provenance record shows the right binding, established without reference answers from XBRL facts and filing rules.

%% file: tables/results_main.tex
\begin{table}[t]
\centering\small
\setlength{\tabcolsep}{2.6pt}
\begin{tabular}{@{}llrrrrr@{}}
\toprule
 & & & \multicolumn{4}{c}{\% of scored claims} \\
\cmidrule(l){4-7}
Filing & Agent & Scored & Value & Bind. & Supp. & RNWA \\
\midrule
\multirow{3}{*}{Stmts} & Claude & 338 & 98.5 & 98.2 & 96.2 & 1.8 \\
 & GPT & 280 & 99.6 & 99.3 & 97.5 & 0.7 \\
 & DeepSeek & 335 & 91.6 & 92.5 & 81.5 & 5.4 \\
\midrule
\multirow{3}{*}{NQDC} & Claude & 693 & 100.0 & 99.0 & 98.6 & 1.0 \\
 & GPT & 680 & 99.1 & 99.7 & 98.7 & 0.3 \\
 & DeepSeek & 741 & 98.4 & 93.7 & 90.4 & 6.2 \\
\midrule
\multirow{3}{*}{13F} & Claude & 423 & 97.9 & 81.8 & 79.4 & 18.2 \\
 & GPT & 387 & 97.7 & 82.7 & 80.4 & 17.3 \\
 & DeepSeek & 492 & 92.9 & 65.4 & 57.3 & 34.3 \\
\midrule
\multirow{3}{*}{All} & Claude & 1{,}454 & 99.0 & 93.8 & 92.4 & 6.2 \\
 & GPT & 1{,}347 & 98.8 & 94.7 & 93.2 & 5.3 \\
 & DeepSeek & 1{,}568 & 95.2 & 84.6 & 78.1 & 14.9 \\
\bottomrule
\end{tabular}
\caption{Results as \% of scored claims. RNWA: right number, wrong account (passes the value check, fails the binding check). Full results: Appendix~\ref{app:results}.}
\label{tab:results}
\end{table}

%% file: sections/limitations.tex
\section*{Limitations}
\label{sec:limitations}

\paragraph{Scorer.}
All agent results come from our automatic scorer. The expert review estimates its false-fail rate at about 20\% for binding and its false-pass rate at about 5\% (Table~\ref{tab:audit-claims}); the expert saw labels from two LLM annotators, and these labels may have anchored some judgments. We fixed the scorer after inspecting Claude's outputs, so Claude's numbers are in-sample.

\paragraph{Value check and tagging.}
The value check ignores scale and sign, so a change stated in the wrong direction with the right magnitude can pass both the value and binding checks (\S\ref{sec:method-verdicts}). In our runs, 20 of 350 value-pass change claims with a stated direction contradict the sign of the reference value, and all 20 fail the binding check. A regular-expression tagger extracts claims, and a missed or misread number can look like an agent error; the share of claims left unscored differs widely across agents (\S\ref{sec:results-design}).

\paragraph{Reference answers.}
The domain review covered 200 questions of an early version. The re-implementation checks that the reference answers follow the written rules, not that the rules read the filings correctly. There is no inter-annotator agreement. Of the 216 rules, 194 are conventions; other reasonable conventions would change some reference values (Appendix~\ref{app:rules}).

{
\paragraph{XBRL evidence.}
Incorrect or overly broad filing tags can affect reference bindings even after domain review.
A missing tag does not establish that a value is undisclosed: the 26 NQDC company-wide slots for which our rules find no value remain \textit{not found by rule}.%
}

\paragraph{Setting.}
All agents run in one four-stage pipeline with fixed retrieval settings and no other tools, so results may not transfer to agents that browse or read more text; we do not test monitors that work without reference answers.

\paragraph{Runs and statistics.}
Each agent answered each question once, and the CLIs do not expose temperature or seed, so we cannot report run-to-run variance. Wilson intervals ignore clustering within questions, and the 13F-GROUP results rest on 9 usable co-filing groups.

\paragraph{Scope and contamination.}
NumTrace covers 30 large U.S. issuers, 30 selected 13F managers, English filings and one frozen snapshot. All filings are public and may be in pre-training data; the support check limits but does not exclude the risk that a model recalls a number without reading the snapshot.

%% file: sections/ethics.tex
\section*{Ethics Statement}
\label{sec:ethics}

\paragraph{Data and licenses.}
NumTrace uses only public filings from SEC EDGAR, which are U.S. public records. Upon publication, we will release the code under the MIT license and the questions, reference answers and frozen snapshot under CC BY 4.0; the license covers our additions, not the filings. Agent trajectories will be released for research use, subject to each provider's terms of use.

\paragraph{Personal information.}
The NQDC questions name executive officers as printed in the companies' Item 402 disclosures, and we add no information about them. The identities of the human reviewers do not appear in the data.

\paragraph{Intended use and risks.}
NumTrace is an evaluation benchmark with no training split. It is not intended for investment, compensation or legal decisions, for training the evaluated models or for ranking companies or executives.

\paragraph{Human participants.}
A financial analyst with eight years of equity research experience reviewed 200 questions of an early version of the reference answers; a programmer re-implemented the rules for 110 questions; and a financial analyst with seven years of equity research experience checked 420 sampled agent outputs against the filings (\S\ref{sec:validation}). All received written instructions and were contacts of the research team; no crowdworkers were used.

%% file: sections/appendix.tex
\appendix

\providecommand{\qid}[1]{{\small\nolinkurl{#1}}}
\makeatletter
\g@addto@macro{\UrlBreaks}{\do\-}
\makeatother

\raggedbottom
\setcounter{topnumber}{4}\setcounter{bottomnumber}{3}\setcounter{totalnumber}{6}
\setcounter{dbltopnumber}{3}
\renewcommand{\topfraction}{0.9}\renewcommand{\bottomfraction}{0.8}
\renewcommand{\dbltopfraction}{0.9}\renewcommand{\textfraction}{0.07}
\renewcommand{\floatpagefraction}{0.8}\renewcommand{\dblfloatpagefraction}{0.8}
\makeatletter
\setlength{\@fptop}{0pt}\setlength{\@fpsep}{12pt}\setlength{\@fpbot}{0pt plus 1fil}
\setlength{\@dblfptop}{0pt}\setlength{\@dblfpsep}{12pt}\setlength{\@dblfpbot}{0pt plus 1fil}
\makeatother

\section{Extended Related Work}
\label{sec:related}
\label{sec:appendix}

Table~\ref{tab:positioning} places NumTrace next to the closest work: what each work uses as reference answers, what unit it scores, whether it checks what a number refers to, whether it locates errors by stage and which filings it uses. We checked every entry against the cited paper. FinDeepIndicator, Fin-RATE and BigFinanceBench also locate errors by stage, so the stage label is not new to NumTrace.

\input{tables/positioning}

\paragraph{Financial evaluation.}
FinBen evaluates financial LLMs across 42 datasets and 24 tasks, including RAG and agent evaluation \citep{xie2024finben}. \citet{lin2025evaluation} present an evaluation and benchmarking suite for financial LLMs and agents across exploration, readiness and governance. Within this broader evaluation agenda, NumTrace audits the source and financial semantics of individual numerical claims.

Numerical QA over filings began with expert-annotated report excerpts \citep{chen2021finqa,zhu2021tatqa,chen2022convfinqa}. Later work moved to full 10-Ks \citep{reddy2024docfinqa} and evidence pages \citep{islam2023financebench}, and two benchmarks take their reference answers from structured data: BizBench from iXBRL tags \citep{krumdick2024bizbench} and SEC-QA from templates over database values \citep{lai2024secqa}. All of them score the final answer.
Fin-RATE lists entity and period mismatches among its error types \citep{jiang2026finrate}. Agentic benchmarks move the focus to search and research \citep{hu2025finsearchcomp,bigeard2025financeagent,wang2026finfirst,zhang2026frontierfinance,yang2026findeepindicator}. FinTrace scores tool-calling trajectories with LLM-judged rubrics and finds that models select tools well but use the retrieved information poorly; it does not check what individual numbers refer to \citep{cao2026fintrace}.

Several recent benchmarks look past the final answer. FinFIRST states the entity, period, definition and unit in each prompt and checks them with atomic rubrics \citep{wang2026finfirst}, BigFinanceBench grades derivations against expert rubrics \citep{wang2026bigfinancebench}, and FinChain checks intermediate reasoning steps \citep{xie2026finchain}. Two systems rank evidence for retrieval: FinCARDS by entity, metric, period and number \citep{zhou2026fincards}, and FinRank by passage \citep{mansouri2026finrank}.
Other work ties numbers to structured facts. VeriFin grounds answer operands in XBRL facts \citep{hall2026verifin}, and FinGround matches the fields of atomic claims against evidence but still returns one label \citep{guo2026finground}. Related systems verify statements against company facts \citep{panda2026finverbench}, extract values with per-value provenance \citep{moslonka2026ledger}, or build on XBRL and other structured SEC data \citep{han2024xbrlagent,wang2025fintagging}.

Most other verification work gives one label per answer or per claim. This holds for work on financial hallucination \citep{kang2023deficiency,zhang2025faith}, search agents \citep{chen2026finsearch} and claim verification \citep{liu2026sameclaim}, for fact checking over tables, numerical claims and financial reports \citep{chen2020tabfact,venktesh2024quantemp,zhao2024findver}, and for general metrics such as SummaC, HaluEval and FActScore \citep{laban2022summac,li2023halueval,min2023factscore}.
Attribution work asks whether a cited passage supports generated text \citep{bohnet2022attributed,gao2023alce,yue2023attrscore}; deep-research evaluations ask the same of each claim in a long report \citep{venkit2025deeptrace,huang2026deepfact}. DeepResearchGym fixes a web snapshot \citep{coelho2025deepresearchgym}, as we fix an EDGAR snapshot.
Agent failure attribution localizes failing steps \citep{zhang2025whowhen,deshpande2025trail}; our first-visible stage is a descriptive label, not a causal attribution.
These works fold the meaning of a number into one rubric item or one label; NumTrace scores the value and the binding of every numerical claim separately.

\paragraph{Financial verification frameworks.}
\citet{guan2026finverify} organize financial-agent verification into five protocols covering temporal integrity, source fidelity, comparability, accounting integrity and financial-model fidelity, with evidence-bound inputs and deterministic checks yielding Accept, Reject or Abstain. NumTrace complements this verification framework by recording value and binding outcomes separately to measure source-faithful but semantically misbound numerical claims.

\paragraph{Financial research applications.}
\citet{anonymous2026memos} validate claims in LLM-generated investment research memos against official disclosures and show that low claim-level refutation rates can coexist with substantial memo-level contamination. This output-level exposure motivates claim-level diagnostics: NumTrace separates value fidelity from financial binding to identify numbers that are copied correctly but used with the wrong financial meaning.

Shell company screening provides a high-value application setting for numerical provenance auditing: ownership stakes, share counts and balance-sheet figures used to assess candidates must refer to the correct entity, date and scope. \citet{lin2026shellscreening} study deep research agents in this setting and identify limitations in information freshness, multi-source reasoning and claim--evidence traceability. These findings motivate extending NumTrace-style audits to screening workflows, which are outside the present benchmark.

{
\paragraph{XBRL and taxonomy-grounded evaluation.}
FinTagging evaluates financial fact extraction and mapping to US-GAAP concepts \citep{wang2025fintagging}.
FinAuditing uses real XBRL filings and data-quality rule messages to construct tasks for financial semantic matching, relationship extraction and mathematical reasoning \citep{wang2025finauditing}.
NumTrace uses tagged facts for statement reference answers and applies the same binding schema to 13F and NQDC; it scores each claim's value and binding separately.%
}

{\color{rnwarevision}\hypersetup{citecolor=rnwarevision,urlcolor=rnwarevision,linkcolor=rnwarevision}
\paragraph{Certified traces.}
\citet{liu2026certified} propose that agents in policy-governed domains execute only traces that a permissibility checker has certified against a policy system, and they note that monitorable reasoning is not the same as certifiability.
NumTrace does not certify traces at run time. On a frozen corpus, it measures the RNWA rate: the share of scored claims that pass the value check but fail the binding check. The provenance record can serve as certification material.

\paragraph{Diagnostic measures of factuality.}
Relation-generation evaluation already checks extracted entity--value--relation tuples against source records \citep{wiseman2017datadocument}.
PARENT aligns generated and reference text with source tables when computing precision and recall \citep{dhingra2019parent}; FActScore measures the proportion of atomic facts supported by a knowledge source \citep{min2023factscore}.
FRANK distinguishes entity, predicate and circumstance errors within a broader typology \citep{pagnoni2021frank}.
NumTrace's diagnostic focus is the joint outcome of source-level value fidelity and financial binding: a claim may pass the former and fail the latter.
The RNWA rate measures that component using recorded source spans, derivations and binding dimensions.%
}

\section{Datasheet}
\label{app:datasheet}
\label{app:release}

We follow \citet{gebru2021datasheets} in short form.

{
\begin{table*}[t]

\centering\small
\setlength{\tabcolsep}{4pt}
\begin{tabular}{@{}>{\raggedright\arraybackslash}p{0.09\textwidth}>{\raggedright\arraybackslash}p{0.22\textwidth}>{\raggedright\arraybackslash}p{0.18\textwidth}>{\raggedright\arraybackslash}p{0.19\textwidth}>{\raggedright\arraybackslash}p{0.22\textwidth}@{}}
\toprule
Field & XBRL counterpart & Statements & Form 13F & NQDC \\
\midrule
Entity & Reporting entity & Issuer CIK & Manager/filer CIK & Issuer; executive identity \\
Concept & Taxonomy concept & \texttt{us-gaap:} & Project \texttt{13f:} & Project \texttt{nqdc:} \\
Period & Instant or duration & Fiscal instant/interval & Report quarter & Fiscal year \\
Unit & Reported unit & USD; USD/shares & USD; shares & USD; percent \\
Scope & Dimensional qualifier (analogy only) & Entity-wide facts & Single filer/co-filing group & Executive/plan/company \\
\bottomrule
\end{tabular}

\caption{Conceptual correspondence between NumTrace binding fields and the XBRL fact model \citep[Section~3.4]{xbrl2021oim}. Scope is a NumTrace field, not an implemented XBRL dimension. Project concept prefixes do not constitute formal XBRL taxonomy extensions.}
\label{tab:xbrl-binding}
\end{table*}
}

\paragraph{Composition.}
Section~\ref{sec:benchmark} gives the composition, and Table~\ref{tab:app-slots} splits the answer slots by status and filing type.
We selected the 30 Form~13F managers from 111 candidates with a fixed rule that favors the traps their filings expose (amendments, co-filing, filings on both sides of the unit switch, multi-row positions), within quotas per manager type.
Questions are drawn by seeded stratified sampling from rule-produced candidate pools, balanced across entities, years and parameters (Table~\ref{tab:app-families}).
13F-UNIT exists only for 2022Q3$\rightarrow$2022Q4, the only quarter pair that straddles the unit switch.

\begin{table}[!htbp]
\centering\footnotesize
\setlength{\tabcolsep}{3pt}
\begin{tabular}{lrrrr}
\toprule
Answer slot status & Stmts & 13F & NQDC & Total \\
\midrule
reported & 314 & 536 & 541 & 1{,}391 \\
derived & 104 & 424 & 133 & 661 \\
text & 0 & 120 & 18 & 138 \\
not reported & 2 & 0 & 82 & 84 \\
not found by rule & 0 & 0 & 26 & 26 \\
not applicable & 0 & 0 & 6 & 6 \\
\midrule
Total & 420 & 1{,}080 & 806 & 2{,}306 \\
\bottomrule
\end{tabular}
\caption{Answer slots by status and filing type. \textit{derived} = computed from reported values by a stated formula; \textit{not reported} = the filing shows there is nothing to report, and the correct answer says so; \textit{not found by rule} = the rules found no number, but nothing in the filing says it is absent (no reference value; scored on binding and support only).}
\label{tab:app-slots}
\end{table}

\begin{table}[!htbp]
\centering\footnotesize
\setlength{\tabcolsep}{3pt}
\begin{tabular}{lrrrrrr}
\toprule
Family & Q & 2022 & 2023 & 2024 & 2025 & 2026 \\
\midrule
ST-DEBT & 9 & 3 & 2 & 2 & 2 & 0 \\
ST-EPS & 60 & 9 & 13 & 13 & 12 & 13 \\
ST-LTD & 12 & 1 & 3 & 3 & 2 & 3 \\
ST-Q4 & 39 & 8 & 11 & 10 & 8 & 2 \\
\midrule
13F-GROUP & 60 & 8 & 13 & 12 & 15 & 12 \\
13F-POS & 60 & 8 & 16 & 13 & 15 & 8 \\
13F-TOP & 60 & 0 & 17 & 17 & 17 & 9 \\
13F-UNIT & 40 & 40 & 0 & 0 & 0 & 0 \\
\midrule
NQDC-402I & 87 & 26 & 19 & 19 & 19 & 4 \\
NQDC-FUND & 18 & 3 & 3 & 5 & 6 & 1 \\
NQDC-LIMIT & 50 & 9 & 14 & 13 & 12 & 2 \\
NQDC-ROLL & 45 & 0 & 11 & 14 & 16 & 4 \\
\midrule
Total & 540 & 115 & 122 & 121 & 124 & 58 \\
\bottomrule
\end{tabular}
\caption{Questions per family and as-of year.}
\label{tab:app-families}
\end{table}

\paragraph{Collection and preprocessing.}
Because filings can be amended, the released snapshot, not a fresh download, is the reference for reproduction.

\begin{figure*}[t]
\centering
\includegraphics[width=\textwidth]{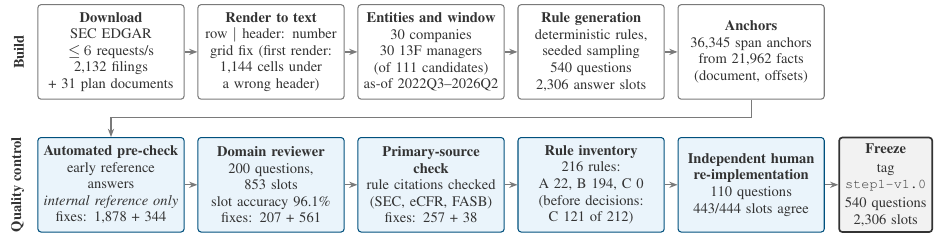}
\caption{Construction and quality control of NumTrace. Top: build steps. Bottom: quality-control rounds; ``fixes'' are lines of the field-level fix logs (statements/13F + NQDC). The automated pre-check only located issues and is not evidence of accuracy. The domain reviewer judged an early version, before its fixes. The rule inventory ends with no unconfirmed rule (class C${}=0$). The independent human re-implementation agrees on 443 of 444 slots.}
\label{fig:construction}
\end{figure*}

\paragraph{Labeling.}
Figure~\ref{fig:construction} shows the build steps and the quality checks.
In the domain review of an early version, slot accuracy was 93.5\% for statements, 98.5\% for 13F and 94.3\% for NQDC; the weakest families were ST-DEBT and ST-LTD.

\paragraph{Known issues.}
Two issues are known. First, 11 of 1,331 reported numeric slots have no anchor (13F 2, NQDC 9); we suggest scoring them on value only. Second, 7 13F-GROUP questions admit an alternative reading (Appendix~\ref{app:rules}).

\section{Rule Inventory and Key Traps}
\label{app:rules}

Every rule that affects reference answers has a class: A (follows an external authority such as the Form~13F instructions and FAQ \citep{sec2023form13f,sec13ffaq} or 17 CFR 229.402 \citep{cfr229402}), B (a design choice confirmed by an author) or C (unconfirmed; none remain).

\begin{itemize}\itemsep 1pt
\item \emph{13F unit by filing date} (A). For 13 manager-quarters, the rule unit disagrees with the unit implied by per-share prices; the reference answer follows the rule.
\item \emph{Co-filing groups} (A). A group total adds the group filer's own rows and each member's rows that the filings tie to the group through the ``Other Managers'' column. No question asks about a group with unclear ties. For 7 questions, one set of rows cannot be tied either way; the reference answer leaves it out and the scorer also accepts the total that includes it.
\item \emph{Item 402(i)} (B). When a printed Total row matches the plan rows within \$1, the reference answer reads that row.
\item \emph{Statements} (B). No question uses a quarter with restated EPS. For \newrev{9 of the 12 long-term-debt (ST-LTD)} questions, the reference answer counts current maturities as long-term debt, which the XBRL definition does not state.
\end{itemize}

\begin{figure}[!htbp]
\centering
\includegraphics[width=\columnwidth]{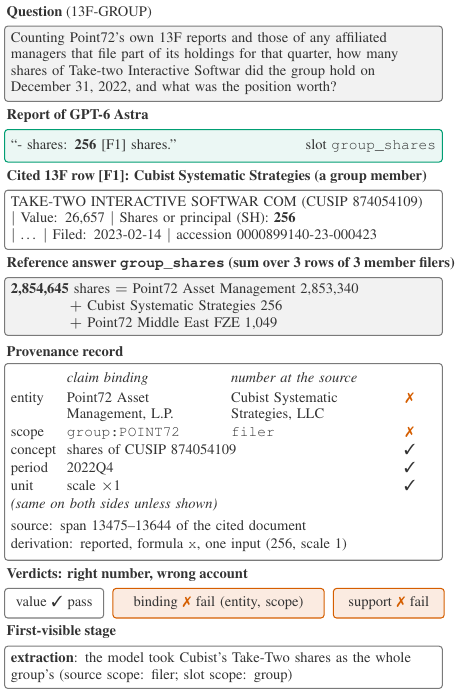}
\caption{Full provenance record of the claim in Figure~\ref{fig:teaser}: a right number, wrong account claim (GPT-6 Astra; question \texttt{S1-13F-GROUP-POINT72-TAKETWO-2022Q4}). The model copies 256 shares exactly from one member filer's 13F row, so value passes. The number at the source belongs to that filer, not to the Point72 group, so binding fails on entity and scope, and support fails against the group total of 2,854,645 shares. The error first becomes visible at extraction. Scorer verdicts, confirmed by the expert review.}
\label{fig:provenance-example}
\end{figure}

\vspace{-12pt}
\section{Prompts and Run Settings}
\label{app:prompts}
\begin{figure*}[tp]
\centering
\includegraphics[width=\textwidth]{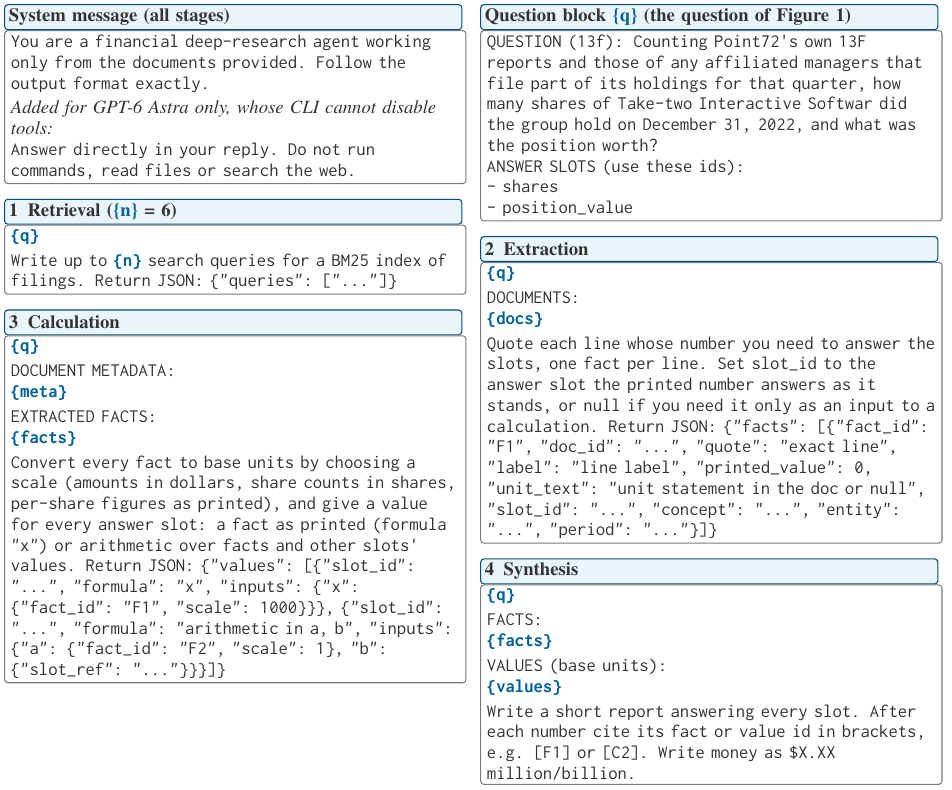}
\caption{The prompts of the four stages, typeset from the harness source code. The harness fills the placeholders in blue.}
\label{fig:prompts}
\end{figure*}

Figure~\ref{fig:prompts} gives the prompts.
BM25 uses \texttt{rank\_bm25} with $k_1=1.5$ and $b=0.75$.
The question block lists only the asked slots, and ids that name a concept or scope are renamed with the question's own words (for example, \texttt{coli\_value} becomes \texttt{life\_insurance\_value}); a unit test checks every prompt for leakage.

Claude Opus~5.5 (\texttt{claude-opus-5-5}) ran through Claude Code CLI 2.1.289 (effort medium) with all tools removed.
GPT-6 Astra (\texttt{gpt-6-astra}) ran through Codex CLI 0.160.0 (effort medium) in a read-only sandbox.
DeepSeek-V4.1-Flash (\texttt{deepseek-flash}) ran as a plain chat completion through DeepSeek's API in thinking mode (effort high); we used the September 10, 2026 release, and all runs were on October 5--6, 2026 \citep{deepseekchangelog}.
The harness logged tool use, and no run used a tool.

\input{tables/audit_claims}
\begin{figure}[t]
  \centering
  \includegraphics[width=\columnwidth]{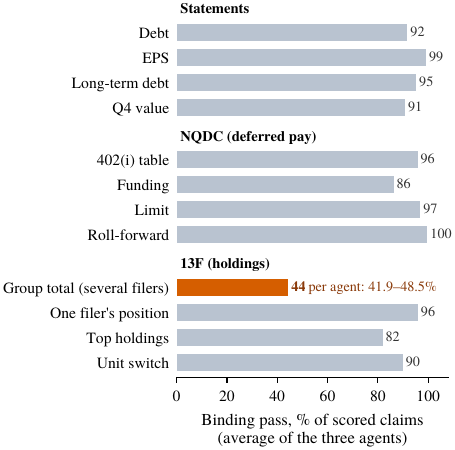}
  \caption{Binding failures concentrate in one family. Each bar averages the binding pass rates of the three agents. Co-filing group totals (orange) are the only family below 80\%: each agent passes 41.9--48.5\%, while one filer's position on the same form reaches 96\%. These rates rest on only 9 usable groups (\S\ref{sec:analysis}).}
  \label{fig:family-binding}
\end{figure}

\section{Scorer Specification}
\label{app:scorer}

\paragraph{Value.}
The tolerance is half a unit of the claim's last stated digit (doubled for a hedge such as ``about'').
A reported claim passes if, for some scale $s \in \{10^{-3}, 1, 10^{3}, 10^{6}, 10^{9}\}$, its magnitude equals the printed number times $s$ within tolerance.
For a computed claim, each input's printed digits must match its source, and the formula is re-evaluated over the agent's own inputs.
Of the RNWA claims, 39, 37 and 103 (Claude, GPT and DeepSeek) are computed claims whose result misses the reference value.

\paragraph{Binding.}
Periods are intervals on each company's fiscal calendar and match only when equal.

\paragraph{Support.}
An anchored reported claim is supported if it passes the value and binding checks; a superseded source fails support.
Multi-row 13F sums are checked row by row against the rows the reference answer sums.

{\color{rnwarevision}\hypersetup{citecolor=rnwarevision,urlcolor=rnwarevision,linkcolor=rnwarevision}
\input{tables/main.tex}
\input{tables/family}
\subsection{Why Marginal Pass Rates Do Not Determine the RNWA Rate}
\label{app:rnwa-marginals}

Write $\overline V=N^{-1}\sum_i V_i$ and $\overline{VB}=N^{-1}\sum_i V_iB_i$.
The identity $1-V_iB_i=(1-V_i)+V_i(1-B_i)$ gives
\begin{equation}
1-\overline{VB}=(1-\overline V)+\widehat r_{\mathrm{RNWA}},
\label{eq:rnwa-decomposition}
\end{equation}
so the RNWA rate is the gap between the value-only and the joint value-and-binding pass rates.

On the same scored set, let $p_{vb}$ be the fraction of claims with $(V,B)=(v,b)$.
The value and binding pass rates are $a=p_{11}+p_{10}$ and $b=p_{11}+p_{01}$, while $\widehat r_{\mathrm{RNWA}}=p_{10}$.
The joint error rate is therefore not generally determined by \mbox{$a$ and $b$:}
\begin{equation}
\max(0,a-b)\leq\widehat r_{\mathrm{RNWA}}
\leq\min(a,1-b).
\label{eq:rnwa-marginal-bounds}
\end{equation}
To see this, write $p_{11}=a-\widehat r_{\mathrm{RNWA}}$, $p_{01}=b-a+\widehat r_{\mathrm{RNWA}}$ and $p_{00}=1-b-\widehat r_{\mathrm{RNWA}}$.
Non-negativity of all four cell proportions gives the bounds; conversely, every rate within them yields a valid joint distribution.

For illustration, consider two hypothetical sets of 100 claims.
One has cell counts $(n_{11},n_{10},n_{01},n_{00})=(80,10,10,0)$; the other has $(90,0,0,10)$.
Both have value and binding pass rates of 90\%, but their RNWA rates are 10\% and 0\%.
These are constructed examples, not experimental results.
Retaining paired verdicts distinguishes whether the two checks fail on the same claims.

The RNWA rate is not simply the difference between the marginal pass rates: $a-b=p_{10}-p_{01}$ includes the reverse mismatch as well.
Nor does a merged pass/fail label identify the RNWA rate: it retains $p_{11}$ but combines $p_{10}$, $p_{01}$ and $p_{00}$ as failures.
No independence assumption is needed.
This argument concerns the scored set and does not identify the RNWA rate among unscored claims.

\paragraph{Conditional RNWA rate and selective risk.}
When $\overline V>0$, the conditional rate $\widehat q=\widehat r_{\mathrm{RNWA}}/\overline V$ is the share of claims passing the value check that fail the binding check.
This has the form of selective risk, with value passing as the acceptance condition and binding failure as the loss \citep{elyaniv2010selective}.
It is an offline diagnostic interpretation, not a guarantee for a deployed checker.
Tables~\ref{tab:results} and~\ref{tab:main} report $\widehat r_{\mathrm{RNWA}}$, whose denominator is all scored claims; neither rate accounts for unscored claims or unanswered slots.%
}

\section{Trace Examples}
\label{app:traces}
Each box below shows one numerical claim from our runs with its provenance record and the three verdicts: two claims that pass every check and four typical failures.
The failures follow the most frequent types: concept errors make up 64\% of RNWA claims (Examples~3 and~4) and period errors 13\% (Example~5). Value errors are rare (105 claims, about 2\% of scored claims); Example~6 shows one caused by a superseded source.
Entity and scope errors (12\%) are illustrated by the Point72 claim in Figure~\ref{fig:provenance-example}.
Examples~1--3 answer the same question, so the passing and failing claims can be compared directly.

\paragraph{How to read a box.}
The agent claim is quoted verbatim; field names such as \texttt{salary\_deferral\_max\_pct} are answer slots from the question block.
In the binding list, a check mark means that the binding the claim states agrees with the source.
For a failed dimension, shown in orange, ``claim'' gives the binding the claim states, and ``cited number'' (or ``inputs'' for a computed claim) gives the binding of the number at its source in the filing.
Concept tags are followed by a plain-language gloss.
Support fails whenever value or binding fails.
The first-visible stage is the first pipeline stage whose logged output shows the error (\S\ref{sec:method-measures}).
Every number, span, binding and verdict is generated from the scored records; only the titles, analyses and glosses are written by hand.

\input{tables/trace_examples}

\section{Scorer Debugging}
\label{app:audit}

Before the scorer fixes described in \S\ref{sec:validation}, 172 of 224 sampled failures were scorer artifacts; after them, 2 of 78 in a new sample were.
The frozen scorer was then applied unchanged to GPT and DeepSeek, whose outputs played no part in developing it.
Two residual artifacts are known: a bare year aligned to the wrong fiscal year, which likely inflates DeepSeek's NQDC period failures, and one 13F filing whose restatement changes the unit.

\section{Additional Results}
\label{app:results}

Table~\ref{tab:main} gives pass rates with Wilson 95\% intervals.
For 13F-GROUP (9 usable groups), resampling groups instead of questions gives binding intervals of [28.6, 68.1] for Claude, [22.0, 63.9] for GPT and [28.8, 58.7] for DeepSeek. Over all 13F questions, a bootstrap gives wider binding intervals than the Wilson intervals in Table~\ref{tab:main}: [75.9, 87.3] for Claude, [76.9, 88.3] for GPT and [58.7, 71.7] for DeepSeek; paired over questions, GPT minus Claude is [$-$3.7, +5.7] points and DeepSeek minus Claude is [$-$22.2, $-$10.4] points.
Table~\ref{tab:family} gives results per family, and Table~\ref{tab:strict} compares lenient and strict reference answers.

\input{tables/strict}

%% file: tables/positioning.tex
\begin{table*}[t]
\centering\footnotesize
\setlength{\tabcolsep}{2.2pt}\definecolor{ntRow}{HTML}{FBEBE1}
\newcommand{\no}{\textcolor{black!45}{No}}\newcommand{\nr}{\textcolor{black!45}{Not stated}}
\begin{tabular}{@{}llll>{\centering\arraybackslash}p{1.1cm}ll@{}}
\toprule
 & Reference & & Fields checked & Separate & & \\
Work & answers & Unit & for a number & verdict & Error stage & Filings \\
\midrule
FinRank & Human labels & Retrieval & \no & \no & \no\ (future work) & 10-K, 10-Q \\
SEC-QA & DB + templates & Answer & \no\ (value within 1\%) & \no & Retrieval vs.\ rest & 10-K, 10-Q, 8-K \\
VeriFin & XBRL facts & Claim & Concept, period, unit & \no & \nr & 10-K, 10-Q \\
Fin-RATE & LLM + humans & Answer & Entity, period, unit & \no & Retrieval vs.\ gen. & Incl.\ proxy \\
FinDeepIndicator & Templates & Task stage & Rows, years & \no & First error & Web data \\
FinFIRST & Expert rubric & Rubric item & Entity, period, unit & \no & Process groups & \nr \\
FinGround & Experts & Claim & Value, unit, period, entity & \no & Retrieval vs.\ reas. & \nr \\
BigFinanceBench & Expert rubric & Rubric item & Entity, period, concept & \no & Rubric stages & Incl.\ proxy \\
FinTrace & LLM + experts & Trajectory & \no & \no & Partly (per turn) & API data \\
FinDVer & Experts & Claim & \no & \no & Partly & 10-K, 10-Q \\
DeepTRACE & LLM judge & Statement & \no & \no & \no & Non-financial \\
DeepFact & Audited experts & Claim & \no & \no & \no & Non-financial \\
\midrule
\rowcolor{ntRow}\textbf{NumTrace} & \textbf{Rules} & \textbf{Claim} & \textbf{All five, incl.\ scope} & \textbf{Yes} & \textbf{Per claim} & 10-K/Q, \textbf{13F, 402(i)} \\
\bottomrule
\end{tabular}
\caption{Positioning against the closest work, checked against each paper. Reference answers: what each work scores against. None uses Form~13F; Fin-RATE and BigFinanceBench use proxy statements but report no Item~402(i) deferred-compensation questions. Fields: entity, concept, period, unit and, for NumTrace, scope. Separate verdict: whether a number's meaning is scored apart from its value. No: not done; Not stated: the paper does not say.}
\label{tab:positioning}
\end{table*}

%% file: tables/audit_claims.tex
\begin{table*}[tp]
\centering\small
\setlength{\tabcolsep}{4pt}
\begin{tabular}{llrrrrrr}
\toprule
& & \multicolumn{2}{c}{False pass} & \multicolumn{2}{c}{False fail} & \multicolumn{2}{c}{Pass rate (weighted)} \\
\cmidrule(lr){3-4}\cmidrule(lr){5-6}\cmidrule(lr){7-8}
Agent & Verdict & sample & weighted \% & sample & weighted \% & expert & scorer \\
\midrule
Claude & value & 1/111 & 0.1 & 6/9 & 75.5 & 99.6 [94.1, 100.0] & 98.9 [92.8, 99.8] \\
 & binding & 5/74 & 0.7 & 11/46 & 17.7 & 94.4 [86.3, 97.8] & 94.0 [85.8, 97.6] \\
 & support & 0/60 & 0.0 & 12/60 & 15.0 & 93.6 [85.3, 97.4] & 92.5 [83.8, 96.7] \\
\midrule
GPT & value & 0/110 & 0.0 & 9/10 & 91.8 & 99.9 [94.6, 100.0] & 99.1 [93.1, 99.9] \\
 & binding & 6/75 & 5.3 & 3/45 & 4.0 & 89.7 [80.3, 94.9] & 94.4 [86.3, 97.9] \\
 & support & 3/60 & 5.0 & 13/60 & 14.1 & 89.5 [80.1, 94.8] & 93.2 [84.6, 97.1] \\
\midrule
DeepSeek & value & 0/108 & 0.0 & 11/12 & 91.5 & 99.6 [95.3, 100.0] & 95.6 [89.3, 98.3] \\
 & binding & 10/76 & 8.7 & 11/44 & 25.1 & 80.7 [71.4, 87.5] & 84.0 [75.1, 90.1] \\
 & support & 6/60 & 9.9 & 17/60 & 28.4 & 76.6 [67.0, 84.2] & 78.2 [68.6, 85.4] \\
\midrule
All & value & 1/329 & 0.0 & 26/31 & 88.9 & 99.7 [97.8, 100.0] & 97.8 [94.9, 99.0] \\
 & binding & 21/225 & 4.9 & 25/135 & 19.8 & 88.0 [83.2, 91.6] & 90.6 [86.1, 93.7] \\
 & support & 9/180 & 4.8 & 42/180 & 23.3 & 86.3 [81.2, 90.1] & 87.6 [82.6, 91.2] \\
\bottomrule
\end{tabular}
\caption{Expert review of 360 sampled claims (60 all-pass and 60 failing per agent). False pass: the scorer passes, the expert judges the claim wrong; false fail: the scorer fails, the expert judges it correct. Weighted figures use inverse inclusion probabilities to estimate rates over all scored claims; pass rates are over all scored claims of the agent, with Wilson 95\% intervals on the effective sample size.}
\label{tab:audit-claims}
\end{table*}

%% file: tables/main.tex
\providecommand{\ci}[2]{\,{\footnotesize[#1,\,#2]}}
\begin{table*}[t]
\centering\small
\setlength{\tabcolsep}{3pt}
\begin{tabular}{llrrlllr}
\toprule
Agent & Filing & Claims & Scored & Value pass & Binding pass & Support pass & RNWA \\
\midrule
\multirow{4}{*}{Claude} & Statements & 529 & 338 & 98.5\ci{96.6}{99.4} & 98.2\ci{96.2}{99.2} & 96.2\ci{93.5}{97.7} & 6 (1.8) \\
 & NQDC & 878 & 693 & 100.0\ci{99.4}{100.0} & 99.0\ci{97.9}{99.5} & 98.6\ci{97.4}{99.2} & 7 (1.0) \\
 & 13F & 894 & 423 & 97.9\ci{96.0}{98.9} & 81.8\ci{77.8}{85.2} & 79.4\ci{75.3}{83.0} & 77 (18.2) \\
\cmidrule(lr){2-8}
 & All & 2{,}301 & 1{,}454 & 99.0\ci{98.4}{99.4} & 93.8\ci{92.5}{94.9} & 92.4\ci{91.0}{93.7} & 90 (6.2) \\
\midrule
\multirow{4}{*}{GPT} & Statements & 282 & 280 & 99.6\ci{98.0}{99.9} & 99.3\ci{97.4}{99.8} & 97.5\ci{94.9}{98.8} & 2 (0.7) \\
 & NQDC & 680 & 680 & 99.1\ci{98.1}{99.6} & 99.7\ci{98.9}{99.9} & 98.7\ci{97.5}{99.3} & 2 (0.3) \\
 & 13F & 445 & 387 & 97.7\ci{95.6}{98.8} & 82.7\ci{78.6}{86.1} & 80.4\ci{76.1}{84.0} & 67 (17.3) \\
\cmidrule(lr){2-8}
 & All & 1{,}407 & 1{,}347 & 98.8\ci{98.1}{99.3} & 94.7\ci{93.4}{95.8} & 93.2\ci{91.7}{94.4} & 71 (5.3) \\
\midrule
\multirow{4}{*}{DeepSeek} & Statements & 377 & 335 & 91.6\ci{88.2}{94.2} & 92.5\ci{89.2}{94.9} & 81.5\ci{77.0}{85.3} & 18 (5.4) \\
 & NQDC & 850 & 741 & 98.4\ci{97.2}{99.1} & 93.7\ci{91.7}{95.2} & 90.4\ci{88.1}{92.3} & 46 (6.2) \\
 & 13F & 752 & 492 & 92.9\ci{90.3}{94.8} & 65.4\ci{61.1}{69.5} & 57.3\ci{52.9}{61.6} & 169 (34.3) \\
\cmidrule(lr){2-8}
 & All & 1{,}979 & 1{,}568 & 95.2\ci{94.0}{96.2} & 84.6\ci{82.7}{86.3} & 78.1\ci{76.0}{80.1} & 233 (14.9) \\
\bottomrule
\end{tabular}
\caption{Pass rates with intervals (Claude Opus~5.5, GPT-6 Astra, DeepSeek-V4.1-Flash; frozen scorer, lenient reference answers). Pass rates (\%) are over scored claims (value verdict not \texttt{na}); brackets are Wilson 95\% intervals. RNWA = right number, wrong account (value pass, binding fail), count and share of scored claims. Each agent answered all 540 questions (120 statements, 200 NQDC, 220 13F).}
\label{tab:main}
\end{table*}

%% file: tables/family.tex
\begin{table*}[t]
\centering\small
\setlength{\tabcolsep}{4pt}
\begin{tabular}{lrrrrrrrrrr}
\toprule
 & & \multicolumn{3}{c}{Claude} & \multicolumn{3}{c}{GPT} & \multicolumn{3}{c}{DeepSeek} \\
\cmidrule(lr){3-5}\cmidrule(lr){6-8}\cmidrule(lr){9-11}
Family & Qs & Scored & Bind.\,\% & RNWA & Scored & Bind.\,\% & RNWA & Scored & Bind.\,\% & RNWA \\
\midrule
ST-DEBT & 9 & 24 & 95.8 & 1 & 16 & 93.8 & 1 & 21 & 85.7 & 3 \\
ST-EPS & 60 & 205 & 99.5 & 1 & 188 & 100.0 & 0 & 213 & 98.1 & 4 \\
ST-LTD & 12 & 37 & 100.0 & 0 & 32 & 100.0 & 0 & 35 & 85.7 & 2 \\
ST-Q4 & 39 & 72 & 94.4 & 4 & 44 & 97.7 & 1 & 66 & 80.3 & 9 \\
\midrule
NQDC-402I & 87 & 298 & 99.3 & 2 & 282 & 100.0 & 0 & 303 & 88.1 & 36 \\
NQDC-FUND & 18 & 6 & 100.0 & 0 & 7 & 100.0 & 0 & 17 & 58.8 & 7 \\
NQDC-LIMIT & 50 & 91 & 95.6 & 4 & 75 & 97.3 & 2 & 87 & 97.7 & 2 \\
NQDC-ROLL & 45 & 298 & 99.7 & 1 & 316 & 100.0 & 0 & 334 & 99.4 & 1 \\
\midrule
13F-GROUP & 60 & 103 & 48.5 & 53 & 86 & 41.9 & 50 & 121 & 43.0 & 69 \\
13F-POS & 60 & 75 & 100.0 & 0 & 64 & 100.0 & 0 & 98 & 87.8 & 12 \\
13F-TOP & 60 & 204 & 89.7 & 21 & 192 & 91.7 & 16 & 215 & 64.2 & 76 \\
13F-UNIT & 40 & 41 & 92.7 & 3 & 45 & 97.8 & 1 & 58 & 79.3 & 12 \\
\bottomrule
\end{tabular}
\caption{Per question family (lenient reference answers). Bind.\,\% = binding pass rate over scored claims; RNWA = count of value-pass, binding-fail claims.}
\label{tab:family}
\end{table*}

%% file: tables/trace_examples.tex
\definecolor{ntClaude}{HTML}{E69F00}
\definecolor{ntGPT}{HTML}{009E73}
\definecolor{ntDeepSeek}{HTML}{0072B2}
\definecolor{ntFail}{HTML}{D55E00}
\definecolor{ntInk}{HTML}{333333}
\definecolor{ntLine}{HTML}{7A7A7A}
\definecolor{ntFillA}{HTML}{F2F2F2}
\definecolor{ntFillB}{HTML}{EAF4FB}
\definecolor{ntFillFail}{HTML}{FBEBE1}
\providecommand{\cmark}{\ding{51}}
\providecommand{\xmark}{\ding{55}}
\DeclareUrlCommand\ntid{\urlstyle{tt}\def\UrlBreaks{\do\-\do\_\do\:\do\[\do\]\do\/\do\.\do\,\do\A\do\B\do\C\do\D\do\E\do\F\do\G\do\H\do\I\do\J\do\K\do\L\do\M\do\N\do\O\do\P\do\Q\do\R\do\S\do\T\do\U\do\V\do\W\do\X\do\Y\do\Z}}
\newenvironment{ntbind}{\begin{list}{}{\setlength{\leftmargin}{3.6em}\setlength{\labelwidth}{3.2em}\setlength{\labelsep}{0.4em}\setlength{\itemsep}{0pt}\setlength{\parsep}{0pt}\setlength{\topsep}{1pt}\renewcommand{\makelabel}[1]{\textit{##1}\hfil}}}{\end{list}}
\newtcolorbox{ntcase}[2]{enhanced, breakable, lines before break=8, colback=white, colframe=ntLine, boxrule=0.5pt, arc=1.5pt, boxsep=1pt,
  left=2pt, right=2pt, top=1pt, bottom=1pt, colbacktitle=white, coltitle=ntInk, fonttitle=\bfseries\small,
  titlerule=0.4pt, titlerule style={ntLine}, title={#1}, title after break={#1 (continued)}, fontupper=\footnotesize\color{ntInk}, before skip=4pt, after skip=5pt}
\newtcolorbox{ntq}{enhanced, breakable, colback=white, colframe=white, boxrule=0pt, borderline={0.5pt}{0pt}{ntLine,dashed},
  arc=1.5pt, boxsep=1pt, left=2pt, right=2pt, top=1pt, bottom=1pt, before skip=2pt, after skip=3pt}
\newtcolorbox{ntclaim}{enhanced, breakable, colback=ntFillA, colframe=ntFillA, boxrule=0pt, arc=1.5pt, boxsep=1pt,
  left=2pt, right=2pt, top=1pt, bottom=1pt, before skip=2pt, after skip=3pt}
\newtcolorbox{ntprov}[1][]{enhanced, colback=ntFillB, colframe=ntFillB, boxrule=0pt, sharp corners, boxsep=0pt,
  left={3.5pt}, right={3.5pt}, top=1pt, bottom=1pt, before skip=0pt, after skip=0pt, #1}
\newtcolorbox{ntanalysis}[1]{enhanced, unbreakable, colback=#1, colframe=#1, boxrule=0pt, arc=1.5pt, boxsep=1pt,
  left=2pt, right=2pt, top=1pt, bottom=1pt, before skip=3pt, after skip=1pt}
\newcommand{\ntfield}[2]{\par\noindent\hangindent=1.2em\textit{#1:} #2\par}
\newcommand{\ntpill}[2]{\tcbox[on line, colback=#1!18, colframe=#1, boxrule=0.4pt, arc=1.5pt, boxsep=0.8pt, left=1.5pt, right=1.5pt, top=0pt, bottom=0pt]{\strut\scriptsize #2}}
\newcommand{\ntchippass}[1]{\tcbox[on line, colback=white, colframe=ntLine, boxrule=0.4pt, arc=1.5pt, boxsep=0.8pt, left=2pt, right=2pt, top=0pt, bottom=0pt]{\strut\scriptsize\textbf{#1}~\cmark}}
\newcommand{\ntchipfail}[1]{\tcbox[on line, colback=ntFillFail, colframe=ntFail, boxrule=0.4pt, arc=1.5pt, boxsep=0.8pt, left=2pt, right=2pt, top=0pt, bottom=0pt]{\strut\scriptsize\textbf{#1}~{\color{ntFail}\xmark}}}
\newcommand{\ntchipna}[1]{\tcbox[on line, colback=white, colframe=ntLine, boxrule=0.4pt, arc=1.5pt, boxsep=0.8pt, left=2pt, right=2pt, top=0pt, bottom=0pt]{\strut\scriptsize\textbf{#1}~n/a}}
\newcommand{\ntverdicts}[1]{\par\smallskip\noindent\textbf{Verdicts.}\ #1\par}

\begin{ntcase}{Example 1: Pass: reported number}{ntLine}
{\scriptsize \ntpill{ntDeepSeek}{DeepSeek} 10-K statements $\cdot$ \ntid{S1-ST-DEBT-XOM-FY2024}}\par\smallskip
\begin{ntq}\textbf{Question.} How much debt did ExxonMobil have at the end of fiscal 2024, and how much of it was due within one year?\end{ntq}
\begin{ntclaim}\textbf{Agent claim.} Of that, \textbf{\$4,955.00 million} (\$4.955 billion) was due within one year [C1][F1].\end{ntclaim}
\begin{ntprov}[top={2.5pt},before skip=2pt]
\textbf{Provenance record.}\par
\ntfield{Span}{\ntid{xom-10k-fy2024-000010-s047 [3944:4003]}\\ ``Total | December 31, 2024: 4,955 | December 31, 2023: 4,090''}
\end{ntprov}
\begin{ntprov}
\ntfield{Binding}{\begin{ntbind}
\item[Entity] \texttt{Exxon Mobil Corporation} \hfill{\cmark}
\item[Concept] \ntid{us-gaap:DebtCurrent} (debt due within one year) \hfill{\cmark}
\item[Period] \ntid{FY2024} \hfill{\cmark}
\item[Unit] \ntid{USD} \hfill{\cmark}
\end{ntbind}}
\end{ntprov}
\begin{ntprov}[bottom={2.5pt},after skip=3pt]
\ntfield{Reference}{4,955 million (10-K, filed 2025-02-19)}
\end{ntprov}
\par\noindent\begin{minipage}{\linewidth}
\ntverdicts{\ntchippass{Value}\ \ntchippass{Binding}\ \ntchippass{Support}}
\begin{ntanalysis}{ntFillA}\textbf{Analysis.} The cited span is the total of notes and loans payable, which is the debt due within one year that the question asks for. The number is printed in the cited 10-K and every binding field matches.\end{ntanalysis}
\end{minipage}
\end{ntcase}

\begin{ntcase}{Example 2: Pass: computed number}{ntLine}
{\scriptsize \ntpill{ntClaude}{Claude} 10-K statements $\cdot$ \ntid{S1-ST-DEBT-XOM-FY2024}}\par\smallskip
\begin{ntq}\textbf{Question.} How much debt did ExxonMobil have at the end of fiscal 2024, and how much of it was due within one year?\end{ntq}
\begin{ntclaim}\textbf{Agent claim.} \textbf{Total debt:} ExxonMobil had total debt of \textbf{\$41.71 billion} at December 31, 2024 [C2].\end{ntclaim}
\begin{ntprov}[top={2.5pt},before skip=2pt]
\textbf{Provenance record.}\par
\ntfield{Input $a$}{\ntid{xom-10k-fy2024-000010-s047 [3944:4003]}\\ ``Total | December 31, 2024: 4,955 | December 31, 2023: 4,090''}
\end{ntprov}
\begin{ntprov}
\ntfield{Input $b$}{\ntid{xom-10k-fy2024-000010-s050 [8410:8486]}\\ ``Total long-term debt | December 31, 2024: 36,755 | December 31, 2023: 37,483''}
\ntfield{Derivation}{$a + b$ over the cited inputs $=$ \$41.71 billion}
\end{ntprov}
\begin{ntprov}
\ntfield{Binding}{\begin{ntbind}
\item[Entity] \texttt{Exxon Mobil Corporation} \hfill{\cmark}
\item[Concept] \ntid{us-gaap:DebtAndCapitalLeaseObligations} (total debt) \hfill{\cmark}
\item[Period] \ntid{FY2024} \hfill{\cmark}
\item[Unit] \ntid{USD} \hfill{\cmark}
\end{ntbind}}
\end{ntprov}
\begin{ntprov}[bottom={2.5pt},after skip=3pt]
\ntfield{Reference}{41,710 million}
\end{ntprov}
\par\noindent\begin{minipage}{\linewidth}
\ntverdicts{\ntchippass{Value}\ \ntchippass{Binding}\ \ntchippass{Support}}
\begin{ntanalysis}{ntFillA}\textbf{Analysis.} Both inputs are copied from cited spans, and re-evaluating $a+b$ over them gives 41,710 million, the reference total. The record lets a reader replay the arithmetic.\end{ntanalysis}
\end{minipage}
\end{ntcase}

\begin{ntcase}{Example 3: Right number, wrong account: concept}{ntFail}
{\scriptsize \ntpill{ntGPT}{GPT} 10-K statements $\cdot$ \ntid{S1-ST-DEBT-XOM-FY2024}}\par\smallskip
\begin{ntq}\textbf{Question.} How much debt did ExxonMobil have at the end of fiscal 2024, and how much of it was due within one year?\end{ntq}
\begin{ntclaim}\textbf{Agent claim.} At the end of fiscal 2024, ExxonMobil had total debt of \textbf{\$41.71 billion [F2]}, including \textbf{\$4.89 billion [F1]} of long-term debt due within one year.\end{ntclaim}
\begin{ntprov}[top={2.5pt},before skip=2pt]
\textbf{Provenance record.}\par
\ntfield{Span}{\ntid{xom-10k-fy2024-000010-s047 [3855:3943]}\\ ``Long-term debt due within one year | December 31, 2024: 4,892 | December 31, 2023: 4,009''}
\end{ntprov}
\begin{ntprov}
\ntfield{Binding}{\begin{ntbind}
\item[Entity] \texttt{Exxon Mobil Corporation} \hfill{\cmark}
\item[Concept] {\color{ntFail}claim: \ntid{us-gaap:DebtCurrent} (debt due within one year); cited number: \ntid{us-gaap:LongTermDebtAndCapitalLeaseObligationsCurrent} (long-term debt due within one year)} \hfill{\color{ntFail}\xmark}
\item[Period] \ntid{FY2024} \hfill{\cmark}
\item[Unit] \ntid{USD} \hfill{\cmark}
\end{ntbind}}
\ntfield{First visible at}{extraction}
\end{ntprov}
\begin{ntprov}[bottom={2.5pt},after skip=3pt]
\ntfield{Reference}{4,955 million (10-K, filed 2025-02-19)}
\end{ntprov}
\par\noindent\begin{minipage}{\linewidth}
\ntverdicts{\ntchippass{Value}\ \ntchipfail{Binding}\ \ntchipfail{Support}}
\begin{ntanalysis}{ntFillFail}\textbf{Analysis.} 4,892 is printed in the cited table, so the value check passes. That row is only the long-term debt due within one year; the debt due within one year is the table total, 4,955 million, which also includes 63 million of bank loans. The claim attaches a correct number to the wrong concept.\end{ntanalysis}
\end{minipage}
\end{ntcase}

\begin{ntcase}{Example 4: Right number, wrong account: concept (computed)}{ntFail}
{\scriptsize \ntpill{ntDeepSeek}{DeepSeek} Form 13F $\cdot$ \ntid{S1-13F-TOP-JEF-TOP3-2025Q3-2025Q4}}\par\smallskip
\begin{ntq}\textbf{Question.} According to its 13F filings, which were Jefferies Financial Group's three largest positions by value on December 31, 2025, and how did the value of each change from September 30, 2025? Count a position as the sum of all of the filer's rows for one CUSIP (every other included manager, shares and principal amounts), leaving out put and call option rows.\end{ntq}
\begin{ntclaim}\textbf{Agent claim.} Change from September 30, 2025: \textbf{+\$243.54 million} [C1], from \$1.73 million [F2] to \$245.26 million [F1].\end{ntclaim}
\begin{ntprov}[top={2.5pt},before skip=2pt]
\textbf{Provenance record.}\par
\ntfield{Input $a$}{\ntid{jef-13f-2025q4-0000096223-26-000010-s001 [10973:11153]}\\ ``ALPHABET INC CAP STK CL C (CUSIP 02079K107) | Value: 245,263,869 | Shares or principal (SH): 781,593 | Investment discretion: DFND | Other managers: \#1 \#5 \#7 \#8 | Filed: 2026-02-09''}
\end{ntprov}
\begin{ntprov}
\ntfield{Input $b$}{\ntid{jef-13f-2025q3-0000096223-25-000015-s001 [7958:8128]}\\ ``ALPHABET INC CAP STK CL C (CUSIP 02079K107) | Value: 1,726,270 | Shares or principal (SH): 7,088 | Investment discretion: DFND | Other managers: \#5 \#8 | Filed: 2025-11-13''}
\ntfield{Derivation}{$a - b$ over the cited inputs $=$ \$243.54 million}
\end{ntprov}
\begin{ntprov}
\ntfield{Binding}{\begin{ntbind}
\item[Entity] \texttt{Jefferies Financial Group Inc.} \hfill{\cmark}
\item[Concept] {\color{ntFail}claim: \ntid{13f:ValueChange[78462F103]} (change in value, SPDR S\&P 500 ETF Trust); inputs: \ntid{13f:Value[02079K107]} (value, Alphabet Class~C)} \hfill{\color{ntFail}\xmark}
\item[Period] \ntid{2025Q3-2025Q4} \hfill{\cmark}
\item[Unit] \ntid{USD} \hfill{\cmark}
\end{ntbind}}
\ntfield{First visible at}{extraction}
\end{ntprov}
\begin{ntprov}[bottom={2.5pt},after skip=3pt]
\ntfield{Reference}{\$70,715,106 $=$ 796,544,244 $-$ 725,829,138 (\ntid{13f:Value[78462F103]} (value, SPDR S\&P 500 ETF Trust))}
\end{ntprov}
\par\noindent\begin{minipage}{\linewidth}
\ntverdicts{\ntchippass{Value}\ \ntchipfail{Binding}\ \ntchipfail{Support}}
\begin{ntanalysis}{ntFillFail}\textbf{Analysis.} Both inputs are copied exactly and $a-b$ is correct, so the value check passes. They are rows for Alphabet Class~C, but the largest position on December 31, 2025 is the SPDR S\&P 500 ETF Trust (CUSIP 78462F103, \$796.5 million), which the agent left out; its change is 796,544,244 $-$ 725,829,138 = 70,715,106. The result is bound to the wrong security.\end{ntanalysis}
\end{minipage}
\end{ntcase}

\begin{ntcase}{Example 5: Right number, wrong account: period}{ntFail}
{\scriptsize \ntpill{ntClaude}{Claude} NQDC (proxy) $\cdot$ \ntid{S1-NQDC-LIMIT-AAPL-MAX-FY2022}}\par\smallskip
\begin{ntq}\textbf{Question.} According to Apple's proxy statement for fiscal 2022, under the Deferred Compensation Plan, up to what percentage of base salary and up to what percentage of annual bonus (cash incentive) could participants elect to defer?\end{ntq}
\begin{ntclaim}\textbf{Agent claim.} \textbf{salary\_deferral\_max\_pct:} Participants could defer up to \textbf{50\%} [C1] of their annual base salary and commissions [F1][F3].\end{ntclaim}
\begin{ntprov}[top={2.5pt},before skip=2pt]
\textbf{Provenance record.}\par
\ntfield{Span}{\ntid{aapl-def14a-fy2024-000008-s017 [5379:5514]}\\ ``Compensation Plan participants may elect to defer up to 50\% of their annual base salary and commissions and up to 90\% of their eligible''}
\end{ntprov}
\begin{ntprov}
\ntfield{Binding}{\begin{ntbind}
\item[Entity] \texttt{Apple Inc.} \hfill{\cmark}
\item[Concept] \ntid{nqdc:SalaryDeferralMaxPct} (maximum salary deferral) \hfill{\cmark}
\item[Period] {\color{ntFail}claim: \ntid{FY2022}; cited number: \ntid{FY2024}} \hfill{\color{ntFail}\xmark}
\item[Unit] \ntid{percent} \hfill{\cmark}
\item[Scope] \texttt{plan:Deferred Compensation Plan} \hfill{\cmark}
\end{ntbind}}
\ntfield{First visible at}{calculation}
\end{ntprov}
\begin{ntprov}[bottom={2.5pt},after skip=3pt]
\ntfield{Reference}{50\% (\ntid{aapl-def14a-fy2022-000019-s018})\\ ``Deferred Compensation Plan participants may elect to defer up to 50\% of their annual base salary and commissions and up to 90\% of their eligible cash bonus, or up to 100\% of their annual cash retainer in the case of Non-Employee Directors, by timely completing a deferral election form.''}
\end{ntprov}
\par\noindent\begin{minipage}{\linewidth}
\ntverdicts{\ntchippass{Value}\ \ntchipfail{Binding}\ \ntchipfail{Support}}
\begin{ntanalysis}{ntFillFail}\textbf{Analysis.} The 50\% limit is copied correctly, but from Apple's fiscal 2024 proxy statement, not the fiscal 2022 statement the question names. The fiscal 2022 statement states the same limit, so a check against the reference value alone would accept the claim; binding fails on period because the evidence describes another fiscal year.\end{ntanalysis}
\end{minipage}
\end{ntcase}

\begin{ntcase}{Example 6: Value error: superseded source}{ntFail}
{\scriptsize \ntpill{ntDeepSeek}{DeepSeek} NQDC (proxy) $\cdot$ \ntid{S1-NQDC-LIMIT-KO-MAX-FY2023}}\par\smallskip
\begin{ntq}\textbf{Question.} According to Coca-Cola's proxy statement for fiscal 2023, under the Deferred Compensation Plan, up to what percentage of base salary and up to what percentage of annual bonus (cash incentive) could participants elect to defer?\end{ntq}
\begin{ntclaim}\textbf{Agent claim.} \textbf{Annual bonus (cash incentive) deferral maximum:} up to \textbf{100\%} of Annual Incentive [F2][C2]\end{ntclaim}
\begin{ntprov}[top={2.5pt},before skip=2pt]
\textbf{Provenance record.}\par
\ntfield{Span}{\ntid{ko-ex1016-001506-a2202147zex1016-s002 [2975:3083]}\\ ``An Eligible Employee may elect to defer up to 80\% of his Base Salary and up to 100\% of his Annual Incentive.''}
\end{ntprov}
\begin{ntprov}
\ntfield{Binding}{\begin{ntbind}
\item[Entity] \texttt{The Coca-Cola Company} \hfill{\cmark}
\item[Concept] \ntid{nqdc:BonusDeferralMaxPct} (maximum bonus deferral) \hfill{\cmark}
\item[Period] \ntid{FY2023} \hfill{\cmark}
\item[Unit] \ntid{percent} \hfill{\cmark}
\item[Scope] \texttt{plan:Deferred Compensation Plan} \hfill{\cmark}
\end{ntbind}}
\ntfield{First visible at}{retrieval}
\end{ntprov}
\begin{ntprov}[bottom={2.5pt},after skip=3pt]
\ntfield{Reference}{95\% (\ntid{ko-def14a-fy2023-003468-s064})\\ ``Eligible participants may defer up to 80\% of their base salary and up to 95\% of their annual incentive.''}
\end{ntprov}
\par\noindent\begin{minipage}{\linewidth}
\ntverdicts{\ntchipfail{Value}\ \ntchippass{Binding}\ \ntchipfail{Support}}
\begin{ntanalysis}{ntFillFail}\textbf{Analysis.} The cited sentence comes from a plan document that a later version superseded, and it allows 100\%. Coca-Cola's fiscal 2023 proxy statement, which the question names, states 95\%, so the value check fails and the source is flagged as superseded.\end{ntanalysis}
\end{minipage}
\end{ntcase}

%% file: tables/strict.tex
\begin{table}[htbp]
\centering\footnotesize
\setlength{\tabcolsep}{2pt}
\begin{tabular}{@{}lrrrr@{}}
\toprule
Filing & Value & Binding & Support & RNWA \\
\midrule
\multicolumn{5}{@{}l}{\textbf{Claude}} \\
Statements & 98.5 / 98.5 & 98.2 / 98.2 & 96.2 / 94.7 & 6 / 6 \\
NQDC & 100.0 / 100.0 & 99.0 / 98.5 & 98.6 / 98.1 & 7 / 10 \\
13F & 97.9 / 97.9 & 81.8 / 81.8 & 79.4 / 79.4 & 77 / 77 \\
\midrule
\multicolumn{5}{@{}l}{\textbf{GPT}} \\
Statements & 99.6 / 99.6 & 99.3 / 99.3 & 97.5 / 96.1 & 2 / 2 \\
NQDC & 99.1 / 99.1 & 99.7 / 99.7 & 98.7 / 98.6 & 2 / 2 \\
13F & 97.7 / 97.7 & 82.7 / 82.7 & 80.4 / 80.4 & 67 / 67 \\
\midrule
\multicolumn{5}{@{}l}{\textbf{DeepSeek}} \\
Statements & 91.6 / 91.6 & 92.5 / 92.5 & 81.5 / 80.9 & 18 / 18 \\
NQDC & 98.4 / 98.4 & 93.7 / 93.3 & 90.4 / 90.0 & 46 / 48 \\
13F & 92.9 / 92.9 & 65.4 / 65.4 & 57.3 / 57.3 & 169 / 169 \\
\bottomrule
\end{tabular}
\caption{Pass rates (\%) and RNWA counts under lenient / strict reference answers (left / right in each cell). Lenient reference answers also accept a slot's listed alternatives and a computed sum of parts for a printed total; strict ones accept only the primary value.}
\label{tab:strict}
\end{table}

%% file: main.bbl
\begin{thebibliography}{70}
\providecommand{\natexlab}[1]{#1}

\bibitem[{Bigeard et~al.(2025)Bigeard, Nashold, Krishnan, and Wu}]{bigeard2025financeagent}
Antoine Bigeard, Langston Nashold, Rayan Krishnan, and Shirley Wu. 2025.
\newblock \href {https://arxiv.org/abs/2508.00828} {Finance agent benchmark: Benchmarking {LLMs} on real-world financial research tasks}.
\newblock \emph{Preprint}, arXiv:2508.00828.

\bibitem[{Bohnet et~al.(2022)Bohnet, Tran, Verga, Aharoni, Andor, Baldini~Soares, Ciaramita, Eisenstein, Ganchev, Herzig, Hui, Kwiatkowski, Ma, Ni, Sestorain~Saralegui, Schuster, Cohen, Collins, Das, Metzler, Petrov, and Webster}]{bohnet2022attributed}
Bernd Bohnet, Vinh~Q. Tran, Pat Verga, Roee Aharoni, Daniel Andor, Livio Baldini~Soares, Massimiliano Ciaramita, Jacob Eisenstein, Kuzman Ganchev, Jonathan Herzig, Kai Hui, Tom Kwiatkowski, Ji~Ma, Jianmo Ni, Lierni Sestorain~Saralegui, Tal Schuster, William~W. Cohen, Michael Collins, Dipanjan Das, and 3 others. 2022.
\newblock \href {https://arxiv.org/abs/2212.08037} {Attributed question answering: Evaluation and modeling for attributed large language models}.
\newblock \emph{Preprint}, arXiv:2212.08037.

\bibitem[{Cao et~al.(2026)Cao, Li, Liu, Cao, Xu, Qian, Peng, Tang, Yao, Huang, Subbalakshmi, Zhu, Suchow, and Yu}]{cao2026fintrace}
Yupeng Cao, Haohang Li, Weijin Liu, Wenbo Cao, Anke Xu, Lingfei Qian, Xueqing Peng, Minxue Tang, Zhiyuan Yao, Jimin Huang, K.~P. Subbalakshmi, Zining Zhu, Jordan~W. Suchow, and Yangyang Yu. 2026.
\newblock \href {https://arxiv.org/abs/2604.10015} {{FinTrace}: Holistic trajectory-level evaluation of {LLM} tool calling for long-horizon financial tasks}.
\newblock \emph{Preprint}, arXiv:2604.10015.

\bibitem[{Chen et~al.(2020)Chen, Wang, Chen, Zhang, Wang, Li, Zhou, and Wang}]{chen2020tabfact}
Wenhu Chen, Hongmin Wang, Jianshu Chen, Yunkai Zhang, Hong Wang, Shiyang Li, Xiyou Zhou, and William~Yang Wang. 2020.
\newblock \href {https://openreview.net/forum?id=rkeJRhNYDH} {{TabFact}: A large-scale dataset for table-based fact verification}.
\newblock In \emph{International Conference on Learning Representations}.

\bibitem[{Chen et~al.(2021)Chen, Chen, Smiley, Shah, Borova, Langdon, Moussa, Beane, Huang, Routledge, and Wang}]{chen2021finqa}
Zhiyu Chen, Wenhu Chen, Charese Smiley, Sameena Shah, Iana Borova, Dylan Langdon, Reema Moussa, Matt Beane, Ting-Hao Huang, Bryan Routledge, and William~Yang Wang. 2021.
\newblock \href {https://doi.org/10.18653/v1/2021.emnlp-main.300} {{F}in{QA}: A dataset of numerical reasoning over financial data}.
\newblock In \emph{Proceedings of the 2021 Conference on Empirical Methods in Natural Language Processing}, pages 3697--3711, Online and Punta Cana, Dominican Republic. Association for Computational Linguistics.

\bibitem[{Chen et~al.(2022)Chen, Li, Smiley, Ma, Shah, and Wang}]{chen2022convfinqa}
Zhiyu Chen, Shiyang Li, Charese Smiley, Zhiqiang Ma, Sameena Shah, and William~Yang Wang. 2022.
\newblock \href {https://doi.org/10.18653/v1/2022.emnlp-main.421} {{C}onv{F}in{QA}: Exploring the chain of numerical reasoning in conversational finance question answering}.
\newblock In \emph{Proceedings of the 2022 Conference on Empirical Methods in Natural Language Processing}, pages 6279--6292, Abu Dhabi, United Arab Emirates. Association for Computational Linguistics.

\bibitem[{Chen et~al.(2026)Chen, Zhu, and Zhang}]{chen2026finsearch}
Zoe Chen, Tianlei Zhu, and Andrew Zhang. 2026.
\newblock \href {https://doi.org/10.1109/IDS69480.2026.00021} {An empirical evaluation of hallucinations of agentic {FinSearch}}.
\newblock In \emph{2026 IEEE 12th International Conference on Intelligent Data and Security (IDS)}, pages 74--78, New York City, NY, USA. IEEE.

\bibitem[{Coelho et~al.(2026)Coelho, Ning, He, Mao, Paladugu, Setlur, Jin, Callan, Magalh{\~a}es, Martins, and Xiong}]{coelho2025deepresearchgym}
Jo{\~a}o Coelho, Jingjie Ning, Jingyuan He, Kangrui Mao, Abhijay~Sai Paladugu, Pranav Setlur, Jiahe Jin, Jamie Callan, Jo{\~a}o Magalh{\~a}es, Bruno Martins, and Chenyan Xiong. 2026.
\newblock \href {https://doi.org/10.1145/3805713.3820409} {{DeepResearchGym}: A free, transparent, and reproducible sandbox for deep research}.
\newblock In \emph{Proceedings of the 2026 International ACM SIGIR Conference on Innovative Concepts and Theories in Information Retrieval}, pages 34--43, Melbourne, Australia. Association for Computing Machinery.

\bibitem[{{DeepSeek}(2026)}]{deepseekchangelog}
{DeepSeek}. 2026.
\newblock Change log.
\newblock DeepSeek API Docs.
\newblock \url{https://api-docs.deepseek.com/updates/}, accessed 2026-10-05.

\bibitem[{{DeepSeek-AI}(2026)}]{deepseek2026v41flash}
{DeepSeek-AI}. 2026.
\newblock \href {https://arxiv.org/abs/2609.19969} {{DeepSeek-V4.1-Flash}: Pushing the limits of {KV} cache compression}.
\newblock \emph{Preprint}, arXiv:2609.19969.

\bibitem[{Deshpande et~al.(2025)Deshpande, Gangal, Mehta, Krishnan, Kannappan, and Qian}]{deshpande2025trail}
Darshan Deshpande, Varun Gangal, Hersh Mehta, Jitin Krishnan, Anand Kannappan, and Rebecca Qian. 2025.
\newblock \href {https://arxiv.org/abs/2505.08638} {{TRAIL}: Trace reasoning and agentic issue localization}.
\newblock \emph{Preprint}, arXiv:2505.08638.

\bibitem[{Dhingra et~al.(2019)Dhingra, Faruqui, Parikh, Chang, Das, and Cohen}]{dhingra2019parent}
Bhuwan Dhingra, Manaal Faruqui, Ankur Parikh, Ming-Wei Chang, Dipanjan Das, and William Cohen. 2019.
\newblock \href {https://doi.org/10.18653/v1/P19-1483} {Handling divergent reference texts when evaluating table-to-text generation}.
\newblock In \emph{Proceedings of the 57th Annual Meeting of the Association for Computational Linguistics}, pages 4884--4895. Association for Computational Linguistics.

\bibitem[{El-Yaniv and Wiener(2010)}]{elyaniv2010selective}
Ran El-Yaniv and Yair Wiener. 2010.
\newblock \href {https://www.jmlr.org/papers/v11/el-yaniv10a.html} {On the foundations of noise-free selective classification}.
\newblock \emph{Journal of Machine Learning Research}, 11:1605--1641.

\bibitem[{Gao et~al.(2023)Gao, Yen, Yu, and Chen}]{gao2023alce}
Tianyu Gao, Howard Yen, Jiatong Yu, and Danqi Chen. 2023.
\newblock \href {https://doi.org/10.18653/v1/2023.emnlp-main.398} {Enabling large language models to generate text with citations}.
\newblock In \emph{Proceedings of the 2023 Conference on Empirical Methods in Natural Language Processing}, pages 6465--6488, Singapore. Association for Computational Linguistics.

\bibitem[{Gebru et~al.(2021)Gebru, Morgenstern, Vecchione, Vaughan, Wallach, Daum{\'e}~III, and Crawford}]{gebru2021datasheets}
Timnit Gebru, Jamie Morgenstern, Briana Vecchione, Jennifer~Wortman Vaughan, Hanna Wallach, Hal Daum{\'e}~III, and Kate Crawford. 2021.
\newblock \href {https://doi.org/10.1145/3458723} {Datasheets for datasets}.
\newblock \emph{Communications of the ACM}, 64(12):86--92.

\bibitem[{Guan et~al.(2026)Guan, Chen, Meng, and Liu}]{guan2026finverify}
Mao Guan, Qian Chen, Ge~Meng, and Xiao-Yang Liu. 2026.
\newblock {FinVerify}: A verification framework for financial agents.
\newblock In \emph{Workshop on Who Verifies the Agents? Toward Reliable Agent Development at the 40th Conference on Neural Information Processing Systems ({NeurIPS} 2026)}.

\bibitem[{Guo et~al.(2026)Guo, Wu, and Yiu}]{guo2026finground}
Dongxin Guo, Jikun Wu, and Siu~Ming Yiu. 2026.
\newblock \href {https://arxiv.org/abs/2604.23588} {{FinGround}: Detecting and grounding financial hallucinations via atomic claim verification}.
\newblock \emph{Preprint}, arXiv:2604.23588.
\newblock Accepted to ACL 2026 Industry Track.

\bibitem[{Hall et~al.(2026)Hall, Shome, and Eiers}]{hall2026verifin}
Bethel Hall, Sachi Shome, and William Eiers. 2026.
\newblock \href {https://arxiv.org/abs/2608.10213} {{VeriFin}: A neurosymbolic framework for verifying {LLM}-generated financial claims}.
\newblock \emph{Preprint}, arXiv:2608.10213.

\bibitem[{Han et~al.(2024)Han, Kang, Jin, Liu, and Yang}]{han2024xbrlagent}
Shijie Han, Haoqiang Kang, Bo~Jin, Xiao-Yang Liu, and Steve~Y. Yang. 2024.
\newblock \href {https://doi.org/10.1145/3677052.3698614} {{XBRL} agent: Leveraging large language models for financial report analysis}.
\newblock In \emph{Proceedings of the 5th ACM International Conference on AI in Finance}, pages 856--864, Brooklyn, NY, USA. Association for Computing Machinery.

\bibitem[{Hu et~al.(2026)Hu, Jiao, Liu, Mutu, Ren, Wen, Zhang, Zhang, Gao, He, Hu, Liao, Wang, Liu, Sun, Zeng, Zeng, Yang, Yang, Yin, Zhang, Zhang, Zhao, Zhu, Namkoong, and Huang}]{hu2025finsearchcomp}
Liang Hu, Jianpeng Jiao, Jiashuo Liu, Dongyuan Mutu, Yanle Ren, Zhoufutu Wen, Kaiyuan Zhang, Xuanliang Zhang, Xiang Gao, Tianci He, Fei Hu, Yali Liao, Zaiyuan Wang, Jingkai Liu, Daibin Sun, Ziqing Zeng, Zhiyuan Zeng, Chenghao Yang, Qianyu Yang, and 7 others. 2026.
\newblock \href {https://openreview.net/forum?id=8AJbbbe2ni} {{FinSearchComp}: Towards a realistic, expert-level evaluation of financial search and reasoning}.
\newblock In \emph{The Fourteenth International Conference on Learning Representations}.

\bibitem[{Huang et~al.(2026)Huang, Ribeiro, Hardalov, Dhingra, Dreyer, and Saligrama}]{huang2026deepfact}
Yukun Huang, Leonardo F.~R. Ribeiro, Momchil Hardalov, Bhuwan Dhingra, Markus Dreyer, and Venkatesh Saligrama. 2026.
\newblock \href {https://doi.org/10.18653/v1/2026.acl-long.1586} {{DeepFact}: Co-evolving benchmarks and agents for deep research factuality}.
\newblock In \emph{Proceedings of the 64th Annual Meeting of the Association for Computational Linguistics (Volume 1: Long Papers)}, pages 34356--34386, San Diego, California, United States. Association for Computational Linguistics.

\bibitem[{{Internal Revenue Service}(2024)}]{irs2024nqdcatg}
{Internal Revenue Service}. 2024.
\newblock Nonqualified deferred compensation audit technique guide.
\newblock Publication 5528 (Rev. 3-2024), Catalog Number 37690C.
\newblock \url{https://www.irs.gov/pub/irs-pdf/p5528.pdf}, accessed 2026-10-05.

\bibitem[{Islam et~al.(2023)Islam, Kannappan, Kiela, Qian, Scherrer, and Vidgen}]{islam2023financebench}
Pranab Islam, Anand Kannappan, Douwe Kiela, Rebecca Qian, Nino Scherrer, and Bertie Vidgen. 2023.
\newblock \href {https://arxiv.org/abs/2311.11944} {{FinanceBench}: A new benchmark for financial question answering}.
\newblock \emph{Preprint}, arXiv:2311.11944.

\bibitem[{Jiang et~al.(2026)Jiang, Chen, Makri, Chen, Li, Maatouk, Tassiulas, Brenner, Xiang, and Ying}]{jiang2026finrate}
Yidong Jiang, Junrong Chen, Eftychia Makri, Jialin Chen, Peiwen Li, Ali Maatouk, Leandros Tassiulas, Eliot Brenner, Bing Xiang, and Rex Ying. 2026.
\newblock \href {https://doi.org/10.1145/3770855.3817528} {{Fin-RATE}: A real-world financial analytics and tracking evaluation benchmark for {LLMs} on {SEC} filings}.
\newblock In \emph{Proceedings of the 32nd ACM SIGKDD Conference on Knowledge Discovery and Data Mining V.2}, pages 9183--9194, Jeju Island, Republic of Korea. Association for Computing Machinery.

\bibitem[{Kang and Liu(2023)}]{kang2023deficiency}
Haoqiang Kang and Xiao-Yang Liu. 2023.
\newblock \href {https://openreview.net/forum?id=SGiQxu8zFL} {Deficiency of large language models in finance: An empirical examination of hallucination}.
\newblock In \emph{I Can't Believe It's Not Better! Workshop: Failure Modes in the Age of Foundation Models at {NeurIPS} 2023}.

\bibitem[{Krumdick et~al.(2024)Krumdick, Koncel-Kedziorski, Lai, Reddy, Lovering, and Tanner}]{krumdick2024bizbench}
Michael Krumdick, Rik Koncel-Kedziorski, Viet~Dac Lai, Varshini Reddy, Charles Lovering, and Chris Tanner. 2024.
\newblock \href {https://doi.org/10.18653/v1/2024.acl-long.452} {{B}iz{B}ench: A quantitative reasoning benchmark for business and finance}.
\newblock In \emph{Proceedings of the 62nd Annual Meeting of the Association for Computational Linguistics (Volume 1: Long Papers)}, pages 8309--8332, Bangkok, Thailand. Association for Computational Linguistics.

\bibitem[{Laban et~al.(2022)Laban, Schnabel, Bennett, and Hearst}]{laban2022summac}
Philippe Laban, Tobias Schnabel, Paul~N. Bennett, and Marti~A. Hearst. 2022.
\newblock \href {https://doi.org/10.1162/tacl_a_00453} {{SummaC}: Re-visiting {NLI}-based models for inconsistency detection in summarization}.
\newblock \emph{Transactions of the Association for Computational Linguistics}, 10:163--177.

\bibitem[{Lai et~al.(2025)Lai, Krumdick, Lovering, Reddy, Schmidt, and Tanner}]{lai2024secqa}
Viet Lai, Michael Krumdick, Charles Lovering, Varshini Reddy, Craig Schmidt, and Chris Tanner. 2025.
\newblock \href {https://doi.org/10.18653/v1/2025.finnlp-2.15} {{SEC-QA}: A systematic evaluation corpus for financial {QA}}.
\newblock In \emph{Proceedings of The 10th Workshop on Financial Technology and Natural Language Processing}, pages 221--236, Suzhou, China. Association for Computational Linguistics.

\bibitem[{Li et~al.(2023)Li, Cheng, Zhao, Nie, and Wen}]{li2023halueval}
Junyi Li, Xiaoxue Cheng, Wayne~Xin Zhao, Jian-Yun Nie, and Ji-Rong Wen. 2023.
\newblock \href {https://doi.org/10.18653/v1/2023.emnlp-main.397} {{HaluEval}: A large-scale hallucination evaluation benchmark for large language models}.
\newblock In \emph{Proceedings of the 2023 Conference on Empirical Methods in Natural Language Processing}, pages 6449--6464, Singapore. Association for Computational Linguistics.

\bibitem[{Lin and Liu(2026)}]{lin2026shellscreening}
Feihan Lin and Xiao-Yang Liu. 2026.
\newblock Deep research agents for shell company screening.
\newblock In \emph{2nd Workshop on Agentic {AI} Benchmarks and Applications for Enterprise Tasks at the 40th Conference on Neural Information Processing Systems ({NeurIPS} 2026)}.

\bibitem[{Lin et~al.(2025)Lin, Patel, Zhang, He, Wang, Wang, White, Xiao, and Liu}]{lin2025evaluation}
Shengyuan Lin, Jaisal Patel, Qinchuan Zhang, Kaiwen He, Keyi Wang, Yan Wang, Matt White, Kairong Xiao, and Xiao-Yang Liu. 2025.
\newblock \href {https://openreview.net/forum?id=sSY4h3MFUB} {Evaluation and benchmarking suite for financial large language models and agents}.
\newblock In \emph{Workshop on Evaluating the Evolving {LLM} Lifecycle: Benchmarks, Emergent Abilities, and Scaling at {NeurIPS} 2025}.

\bibitem[{Liu et~al.(2023)Liu, Wang, Yang, and Zha}]{liu2023fingpt}
Xiao-Yang Liu, Guoxuan Wang, Hongyang Yang, and Daochen Zha. 2023.
\newblock \href {https://openreview.net/forum?id=5BqWC1Fz8F} {Data-centric {FinGPT}: Democratizing internet-scale data for financial large language models}.
\newblock In \emph{Workshop on Instruction Tuning and Instruction Following at {NeurIPS} 2023}, New Orleans, Louisiana, United States.

\bibitem[{Liu et~al.(2026{\natexlab{a}})Liu, Wang, and Capponi}]{liu2026certified}
Xiao-Yang Liu, Xiaodong Wang, and Agostino Capponi. 2026{\natexlab{a}}.
\newblock \href {https://arxiv.org/abs/2605.24462} {No certificate, no execution: Certified traces as a foundation for trustworthy {AI} agents}.
\newblock \emph{Preprint}, arXiv:2605.24462.

\bibitem[{Liu et~al.(2026{\natexlab{b}})Liu, Cao, Jiang, Kabir, Giannouris, Xu, Xu, Zhu, Tariquzzaman, Papadopoulos, Wang, Qian, Peng, Xie, Yuan, Almheiri, Alnajjar, Chen, Stuart, Thompson, Tiwari, Lopez-Lira, Liu, Huang, and Ananiadou}]{liu2026sameclaim}
Zhiwei Liu, Yupeng Cao, Yuechen Jiang, Mohsinul Kabir, Polydoros Giannouris, Chen Xu, Ziyang Xu, Tianlei Zhu, Md. Tariquzzaman, Triantafillos Papadopoulos, Yan Wang, Lingfei Qian, Xueqing Peng, Zhuohan Xie, Ye~Yuan, Saeed Almheiri, Abdulrazzaq Alnajjar, Ming-Bin Chen, Harry Stuart, and 6 others. 2026{\natexlab{b}}.
\newblock \href {https://doi.org/10.18653/v1/2026.findings-acl.479} {Same claim, different judgment: Benchmarking scenario-induced bias in multilingual financial misinformation detection}.
\newblock In \emph{Findings of the Association for Computational Linguistics: {ACL} 2026}, pages 9838--9864, San Diego, California, United States. Association for Computational Linguistics.

\bibitem[{Liu~Yanglet et~al.(2025)Liu~Yanglet, Cao, and Deng}]{liu2025mffms}
Xiao-Yang Liu~Yanglet, Yupeng Cao, and Li~Deng. 2025.
\newblock \href {https://doi.org/10.48550/arXiv.2506.01973} {Multimodal financial foundation models ({MFFMs}): Progress, prospects, and challenges}.
\newblock \emph{Preprint}, arXiv:2506.01973.

\bibitem[{Luo et~al.(2026)Luo, Gong, Yu, and Liu}]{anonymous2026memos}
Chenri Luo, Xukang Gong, Binyan Yu, and Xiao-Yang Liu. 2026.
\newblock {LLM}-generated investment research memos need validation.
\newblock In \emph{Workshop on {TAE} ({Trust-AI-Eval}): Can We Trust {AI} Evaluation? at the 40th Conference on Neural Information Processing Systems ({NeurIPS} 2026)}.

\bibitem[{Mansouri et~al.(2026)Mansouri, Saad, Wahrenburg, Weissel, and Woebbeking}]{mansouri2026finrank}
Sasan Mansouri, Daniel Saad, Mark Wahrenburg, Manu Weissel, and Fabian Woebbeking. 2026.
\newblock \href {https://arxiv.org/abs/2608.07400} {{FinRank}: An evidence-grounded benchmark for financial question answering and retrieval over {SEC} filings}.
\newblock \emph{Preprint}, arXiv:2608.07400.

\bibitem[{Min et~al.(2023)Min, Krishna, Lyu, Lewis, Yih, Koh, Iyyer, Zettlemoyer, and Hajishirzi}]{min2023factscore}
Sewon Min, Kalpesh Krishna, Xinxi Lyu, Mike Lewis, Wen-tau Yih, Pang Koh, Mohit Iyyer, Luke Zettlemoyer, and Hannaneh Hajishirzi. 2023.
\newblock \href {https://doi.org/10.18653/v1/2023.emnlp-main.741} {{FA}ct{S}core: Fine-grained atomic evaluation of factual precision in long form text generation}.
\newblock In \emph{Proceedings of the 2023 Conference on Empirical Methods in Natural Language Processing}, pages 12076--12100, Singapore. Association for Computational Linguistics.

\bibitem[{Moslonka et~al.(2026)Moslonka, de~Vitry, Garnier, Randrianarivo, and Malherbe}]{moslonka2026ledger}
Charles Moslonka, Amaury de~Vitry, Arthur Garnier, Hicham Randrianarivo, and Emmanuel Malherbe. 2026.
\newblock \href {https://arxiv.org/abs/2606.13100} {{LEDGER}: A long-context benchmark of corporate annual reports for grounded financial retrieval and extraction}.
\newblock \emph{Preprint}, arXiv:2606.13100.

\bibitem[{Pagnoni et~al.(2021)Pagnoni, Balachandran, and Tsvetkov}]{pagnoni2021frank}
Artidoro Pagnoni, Vidhisha Balachandran, and Yulia Tsvetkov. 2021.
\newblock \href {https://doi.org/10.18653/v1/2021.naacl-main.383} {Understanding factuality in abstractive summarization with {FRANK}: A benchmark for factuality metrics}.
\newblock In \emph{Proceedings of the 2021 Conference of the North American Chapter of the Association for Computational Linguistics: Human Language Technologies}, pages 4812--4829. Association for Computational Linguistics.

\bibitem[{Panda(2026)}]{panda2026finverbench}
Silu Panda. 2026.
\newblock \href {https://arxiv.org/abs/2605.29586} {{FinVerBench}: Benchmark validity and calibration in large language model financial statement verification}.
\newblock \emph{Preprint}, arXiv:2605.29586.

\bibitem[{Reddy et~al.(2024)Reddy, Koncel-Kedziorski, Lai, Krumdick, Lovering, and Tanner}]{reddy2024docfinqa}
Varshini Reddy, Rik Koncel-Kedziorski, Viet~Dac Lai, Michael Krumdick, Charles Lovering, and Chris Tanner. 2024.
\newblock \href {https://doi.org/10.18653/v1/2024.acl-short.42} {{DocFinQA}: A long-context financial reasoning dataset}.
\newblock In \emph{Proceedings of the 62nd Annual Meeting of the Association for Computational Linguistics (Volume 2: Short Papers)}, pages 445--458, Bangkok, Thailand. Association for Computational Linguistics.

\bibitem[{Robertson and Zaragoza(2009)}]{robertson2009bm25}
Stephen Robertson and Hugo Zaragoza. 2009.
\newblock \href {https://doi.org/10.1561/1500000019} {The probabilistic relevance framework: {BM25} and beyond}.
\newblock \emph{Foundations and Trends in Information Retrieval}, 3(4):333--389.

\bibitem[{Tian et~al.(2024)Tian, Byadgi, Kim, Zha, White, Xiao, and Liu}]{tian2024fingptsearch}
Felix Tian, Ajay Byadgi, Daniel~S. Kim, Daochen Zha, Matt White, Kairong Xiao, and Xiao-Yang Liu. 2024.
\newblock \href {https://doi.org/10.1145/3677052.3698637} {Customized {FinGPT} search agents using foundation models}.
\newblock In \emph{Proceedings of the 5th ACM International Conference on AI in Finance}, pages 469--477, Brooklyn, NY, USA. Association for Computing Machinery.

\bibitem[{{U.S. Securities and Exchange Commission}(2023)}]{sec2023form13f}
{U.S. Securities and Exchange Commission}. 2023.
\newblock Form 13f: Information required of institutional investment managers pursuant to section 13(f) of the securities exchange act of 1934 and rules thereunder (general and special instructions).
\newblock SEC 1685 (07-23), OMB No. 3235-0006.
\newblock \url{https://www.sec.gov/files/form13f.pdf}, accessed 2026-10-05.

\bibitem[{{U.S. Securities and Exchange Commission}(2025)}]{sec2025edgarapis}
{U.S. Securities and Exchange Commission}. 2025.
\newblock \href {https://www.sec.gov/search-filings/edgar-application-programming-interfaces} {{EDGAR} application programming interfaces ({APIs})}.
\newblock Updated April 8, 2025.

\bibitem[{{U.S. Securities and Exchange Commission}(2026{\natexlab{a}})}]{cfr229402}
{U.S. Securities and Exchange Commission}. 2026{\natexlab{a}}.
\newblock 17 {CFR} {\S}~229.402 ({Item} 402 of {Regulation S-K}): Executive compensation.
\newblock Code of Federal Regulations, Title 17, Part 229.
\newblock \url{https://www.ecfr.gov/current/title-17/chapter-II/part-229/subpart-229.400/section-229.402}, accessed 2026-10-05.

\bibitem[{{U.S. Securities and Exchange Commission}(2026{\natexlab{b}})}]{sec2026ecd}
{U.S. Securities and Exchange Commission}. 2026{\natexlab{b}}.
\newblock \href {https://xbrl.sec.gov/ecd/2026/ecd-taxonomy-guide-2026-03-16.pdf} {Executive compensation disclosure taxonomy guide}.
\newblock March 2026.

\bibitem[{{U.S. Securities and Exchange Commission}(2026{\natexlab{c}})}]{sec13fdatasets}
{U.S. Securities and Exchange Commission}. 2026{\natexlab{c}}.
\newblock Form 13f data sets.
\newblock SEC Data \& Research, last updated August 31, 2026.
\newblock \url{https://www.sec.gov/data-research/sec-markets-data/form-13f-data-sets}, accessed 2026-10-05.

\bibitem[{{U.S. Securities and Exchange Commission, Division of Corporation Finance}(2008)}]{seccdi12503}
{U.S. Securities and Exchange Commission, Division of Corporation Finance}. 2008.
\newblock Regulation {S-K} compliance and disclosure interpretations, question 125.03 (item 402(i)).
\newblock Staff guidance, issued July 3, 2008.
\newblock \url{https://www.sec.gov/rules-regulations/staff-guidance/compliance-disclosure-interpretations/regulation-s-k}, accessed 2026-10-05.

\bibitem[{{U.S. Securities and Exchange Commission, Division of Investment Management}(2026)}]{sec13ffaq}
{U.S. Securities and Exchange Commission, Division of Investment Management}. 2026.
\newblock Frequently asked questions about form 13f.
\newblock Staff guidance, last reviewed or updated March 13, 2026.
\newblock \url{https://www.sec.gov/rules-regulations/staff-guidance/division-investment-management-frequently-asked-questions/frequently-asked-questions-about-form-13f}, accessed 2026-10-05.

\bibitem[{V et~al.(2024)V, Anand, Anand, and Setty}]{venktesh2024quantemp}
Venktesh V, Abhijit Anand, Avishek Anand, and Vinay Setty. 2024.
\newblock \href {https://doi.org/10.1145/3626772.3657874} {{QuanTemp}: A real-world open-domain benchmark for fact-checking numerical claims}.
\newblock In \emph{Proceedings of the 47th International ACM SIGIR Conference on Research and Development in Information Retrieval}, pages 650--660, Washington DC, USA. Association for Computing Machinery.

\bibitem[{Venkit et~al.(2026)Venkit, Laban, Zhou, Huang, Mao, and Wu}]{venkit2025deeptrace}
Pranav~Narayanan Venkit, Philippe Laban, Yilun Zhou, Kung-Hsiang Huang, Yixin Mao, and Chien-Sheng Wu. 2026.
\newblock \href {https://openreview.net/forum?id=QkaeTea16Y} {{DeepTRACE}: Auditing deep research {AI} systems for tracking reliability across citations and evidence}.
\newblock In \emph{The Fourteenth International Conference on Learning Representations}.

\bibitem[{Wang et~al.(2026{\natexlab{a}})Wang, Meinhardt, Katz, Kim, Chaudhary, Blagden, and Xu}]{wang2026bigfinancebench}
Alex Wang, Georg Meinhardt, Jacob Katz, Joseph~H. Kim, Pratyush~K. Chaudhary, Chase Blagden, and Eric Xu. 2026{\natexlab{a}}.
\newblock \href {https://arxiv.org/abs/2606.03829} {{BigFinanceBench}: A workflow-grounded benchmark for financial-research agents}.
\newblock \emph{Preprint}, arXiv:2606.03829.

\bibitem[{Wang et~al.(2026{\natexlab{b}})Wang, Xiang, Zhao, Yin, Zhong, Huan, Zhou, Zhu, Zhang, Chang, and Zhou}]{wang2026finfirst}
Wenqing Wang, Haitao Xiang, Xinyi Zhao, Mingming Yin, Ying Zhong, Zhaoxin Huan, Qiheng Zhou, Jin Zhu, Xiaolu Zhang, Shi Chang, and Jun Zhou. 2026{\natexlab{b}}.
\newblock \href {https://arxiv.org/abs/2609.25192} {{FinFIRST}: Benchmarking search agents for financial information retrieval, sourcing and traceability}.
\newblock \emph{Preprint}, arXiv:2609.25192.

\bibitem[{Wang et~al.(2026{\natexlab{c}})Wang, Qian, Peng, Ren, Wang, Han, Feng, Mo, Lin, Zhang, He, Luo, Chen, Wu, Xu, Xu, Huang, Xiong, Liu, Xie, and Nie}]{wang2025fintagging}
Yan Wang, Lingfei Qian, Xueqing Peng, Yang Ren, Keyi Wang, Yi~Han, Dongji Feng, Fengran Mo, Shengyuan Lin, Qinchuan Zhang, Kaiwen He, Chenri Luo, Jianxing Chen, Junwei Wu, Chen Xu, Ziyang Xu, Jimin Huang, Guojun Xiong, Xiao-Yang Liu, and 2 others. 2026{\natexlab{c}}.
\newblock \href {https://arxiv.org/abs/2505.20650} {{FinTagging}: Benchmarking {LLMs} for extracting and structuring financial information}.
\newblock \emph{Preprint}, arXiv:2505.20650.

\bibitem[{Wang et~al.(2026{\natexlab{d}})Wang, Wang, Yang, Patel, Zhao, Mo, Peng, Qian, Chen, Guti{\'e}rrez-Basulto, Huang, Xiong, Liu, and Nie}]{wang2025finauditing}
Yan Wang, Keyi Wang, Shanshan Yang, Jaisal Patel, Jeff Zhao, Fengran Mo, Xueqing Peng, Lingfei Qian, Yankai Chen, V{\'i}ctor Guti{\'e}rrez-Basulto, Jimin Huang, Guojun Xiong, Xiao-Yang Liu, and Jian-Yun Nie. 2026{\natexlab{d}}.
\newblock \href {https://doi.org/10.1145/3805712.3808578} {{FinAuditing}: A financial taxonomy-structured multi-document benchmark for evaluating {LLMs}}.
\newblock In \emph{Proceedings of the 49th International ACM SIGIR Conference on Research and Development in Information Retrieval}, pages 3456--3463, Melbourne, Australia. Association for Computing Machinery.

\bibitem[{Wiseman et~al.(2017)Wiseman, Shieber, and Rush}]{wiseman2017datadocument}
Sam Wiseman, Stuart~M. Shieber, and Alexander~M. Rush. 2017.
\newblock \href {https://doi.org/10.18653/v1/D17-1239} {Challenges in data-to-document generation}.
\newblock In \emph{Proceedings of the 2017 Conference on Empirical Methods in Natural Language Processing}, pages 2253--2263. Association for Computational Linguistics.

\bibitem[{{XBRL International}(2021)}]{xbrl2021oim}
{XBRL International}. 2021.
\newblock \href {https://www.xbrl.org/Specification/oim/REC-2021-10-13/oim-REC-2021-10-13.html} {{Open Information Model 1.0}}.
\newblock Recommendation, 13 October 2021.

\bibitem[{Xie et~al.(2024)Xie, Han, Chen, Xiang, Zhang, He, Xiao, Li, Dai, Feng, Xu, Kang, Kuang, Yuan, Yang, Luo, Zhang, Liu, Xiong, Deng, Jiang, Yao, Li, Yu, Hu, Huang, Liu, Lopez-Lira, Wang, Lai, Wang, Peng, Ananiadou, and Huang}]{xie2024finben}
Qianqian Xie, Weiguang Han, Zhengyu Chen, Ruoyu Xiang, Xiao Zhang, Yueru He, Mengxi Xiao, Dong Li, Yongfu Dai, Duanyu Feng, Yijing Xu, Haoqiang Kang, Ziyan Kuang, Chenhan Yuan, Kailai Yang, Zheheng Luo, Tianlin Zhang, Zhiwei Liu, Guojun Xiong, and 15 others. 2024.
\newblock \href {https://doi.org/10.52202/079017-3033} {{FinBen}: A holistic financial benchmark for large language models}.
\newblock In \emph{Advances in Neural Information Processing Systems}, volume~37, pages 95716--95743.
\newblock Datasets and Benchmarks Track.

\bibitem[{Xie et~al.(2026)Xie, Orel, Thareja, Sahnan, Madmoun, Zhang, Banerjee, Georgiev, Peng, Qian, Huang, Su, Singh, Xing, Elbadry, Xu, Li, Koto, Koychev, Chakraborty, Wang, Lahlou, Stoyanov, Ananiadou, and Nakov}]{xie2026finchain}
Zhuohan Xie, Daniil Orel, Rushil Thareja, Dhruv Sahnan, Hachem Madmoun, Fan Zhang, Debopriyo Banerjee, Georgi~Nenkov Georgiev, Xueqing Peng, Lingfei Qian, Jimin Huang, Jinyan Su, Aaryamonvikram Singh, Rui Xing, Rania Elbadry, Chen Xu, Haonan Li, Fajri Koto, Ivan Koychev, and 6 others. 2026.
\newblock \href {https://doi.org/10.18653/v1/2026.acl-long.662} {{FinChain}: A symbolic benchmark for verifiable chain-of-thought financial reasoning}.
\newblock In \emph{Proceedings of the 64th Annual Meeting of the Association for Computational Linguistics (Volume 1: Long Papers)}, pages 14529--14553, San Diego, California, United States. Association for Computational Linguistics.

\bibitem[{Yang et~al.(2026)Yang, Zhu, Lin, Bai, Liu, Huang, Zimmermann, and Chua}]{yang2026findeepindicator}
Chaoqun Yang, Fengbin Zhu, Xinyu Lin, Long Bai, Xiaoluan Liu, Ke-Wei Huang, Roger Zimmermann, and Tat-Seng Chua. 2026.
\newblock \href {https://arxiv.org/abs/2608.00764} {{FinDeepIndicator}: Benchmarking deep research agents in end-to-end financial indicator construction}.
\newblock \emph{Preprint}, arXiv:2608.00764.

\bibitem[{Yue et~al.(2023)Yue, Wang, Chen, Zhang, Su, and Sun}]{yue2023attrscore}
Xiang Yue, Boshi Wang, Ziru Chen, Kai Zhang, Yu~Su, and Huan Sun. 2023.
\newblock \href {https://doi.org/10.18653/v1/2023.findings-emnlp.307} {Automatic evaluation of attribution by large language models}.
\newblock In \emph{Findings of the Association for Computational Linguistics: EMNLP 2023}, pages 4615--4635, Singapore. Association for Computational Linguistics.

\bibitem[{Zhang et~al.(2025{\natexlab{a}})Zhang, Fu, Warrier, Wang, Tan, and Huang}]{zhang2025faith}
Mengao Zhang, Jiayu Fu, Tanya Warrier, Yuwen Wang, Tianhui Tan, and Ke-wei Huang. 2025{\natexlab{a}}.
\newblock \href {https://doi.org/10.1145/3768292.3770433} {{FAITH}: A framework for assessing intrinsic tabular hallucinations in finance}.
\newblock In \emph{Proceedings of the 6th ACM International Conference on AI in Finance}, pages 159--167, Singapore. Association for Computing Machinery.

\bibitem[{Zhang et~al.(2025{\natexlab{b}})Zhang, Yin, Zhang, Liu, Han, Zhang, Li, Wang, Wang, Chen, and Wu}]{zhang2025whowhen}
Shaokun Zhang, Ming Yin, Jieyu Zhang, Jiale Liu, Zhiguang Han, Jingyang Zhang, Beibin Li, Chi Wang, Huazheng Wang, Yiran Chen, and Qingyun Wu. 2025{\natexlab{b}}.
\newblock \href {https://proceedings.mlr.press/v267/zhang25cq.html} {Which agent causes task failures and when? on automated failure attribution of {LLM} multi-agent systems}.
\newblock In \emph{Proceedings of the 42nd International Conference on Machine Learning}, volume 267 of \emph{Proceedings of Machine Learning Research}, pages 76583--76599. PMLR.

\bibitem[{Zhang et~al.(2026)Zhang, Koyluoglu, Venkatesh, Diehl~Martinez, Bhatia, Alidoust, and Paranjape}]{zhang2026frontierfinance}
Yuhao Zhang, O.~Ozan Koyluoglu, Thejas Venkatesh, Richard Diehl~Martinez, Vishank Bhatia, Arash Alidoust, and Ashwin Paranjape. 2026.
\newblock \href {https://arxiv.org/abs/2608.11683} {{FrontierFinance}: A challenging benchmark for measuring frontier intelligence of finance agents}.
\newblock \emph{Preprint}, arXiv:2608.11683.

\bibitem[{Zhao et~al.(2024)Zhao, Long, Jiang, Wang, Chen, Liu, Tang, Zhang, Zhao, and Cohan}]{zhao2024findver}
Yilun Zhao, Yitao Long, Tintin Jiang, Chengye Wang, Weiyuan Chen, Hongjun Liu, Xiangru Tang, Yiming Zhang, Chen Zhao, and Arman Cohan. 2024.
\newblock \href {https://doi.org/10.18653/v1/2024.emnlp-main.818} {{F}in{DV}er: Explainable claim verification over long and hybrid-content financial documents}.
\newblock In \emph{Proceedings of the 2024 Conference on Empirical Methods in Natural Language Processing}, pages 14739--14752, Miami, Florida, USA. Association for Computational Linguistics.

\bibitem[{Zhou et~al.(2026)Zhou, Zhang, Chen, Zhang, Nakov, and Xie}]{zhou2026fincards}
Yixi Zhou, Fan Zhang, Yu~Chen, Haipeng Zhang, Preslav Nakov, and Zhuohan Xie. 2026.
\newblock \href {https://aclanthology.org/2026.findings-acl.1244/} {{FinCARDS}: Card-based analyst reranking for financial document question answering}.
\newblock In \emph{Findings of the Association for Computational Linguistics: ACL 2026}, pages 24836--24852.

\bibitem[{Zhu et~al.(2021)Zhu, Lei, Huang, Wang, Zhang, Lv, Feng, and Chua}]{zhu2021tatqa}
Fengbin Zhu, Wenqiang Lei, Youcheng Huang, Chao Wang, Shuo Zhang, Jiancheng Lv, Fuli Feng, and Tat-Seng Chua. 2021.
\newblock \href {https://doi.org/10.18653/v1/2021.acl-long.254} {{TAT}-{QA}: A question answering benchmark on a hybrid of tabular and textual content in finance}.
\newblock In \emph{Proceedings of the 59th Annual Meeting of the Association for Computational Linguistics and the 11th International Joint Conference on Natural Language Processing (Volume 1: Long Papers)}, pages 3277--3287, Online. Association for Computational Linguistics.

\bibitem[{Zhu et~al.(2025)Zhu, Ng, Liu, Liu, Zeng, Wang, Tan, Yao, Shao, Xu, Wang, Wang, Lin, Li, Zhu, Zhang, Wang, Feng, Hong, Luan, Huang, and Chua}]{zhu2025findeepresearch}
Fengbin Zhu, Xiang~Yao Ng, Ziyang Liu, Chang Liu, Xianwei Zeng, Chao Wang, Tianhui Tan, Xuan Yao, Pengyang Shao, Min Xu, Zixuan Wang, Jing Wang, Xin Lin, Junfeng Li, Jingxian Zhu, Yang Zhang, Wenjie Wang, Fuli Feng, Richang Hong, and 3 others. 2025.
\newblock \href {https://arxiv.org/abs/2510.13936} {{FinDeepResearch}: Evaluating deep research agents in rigorous financial analysis}.
\newblock \emph{Preprint}, arXiv:2510.13936.

\end{thebibliography}
